\documentclass[10pt,twocolumn,letterpaper]{article}

\usepackage{makecell} 
\usepackage{float}
\usepackage{multirow}
\usepackage{pifont}  
\usepackage{stfloats}

\usepackage[pagenumbers]{cvpr} 
\usepackage{color}

\definecolor{cvprblue}{rgb}{0.21,0.49,0.74}
\usepackage[pagebackref,breaklinks,colorlinks,allcolors=cvprblue]{hyperref}

\def\paperID{6596} 
\def\confName{CVPR}
\def\confYear{2025}

\title{SIDA: Social Media Image Deepfake Detection, Localization and Explanation with Large Multimodal Model}
\author{
     Zhenglin Huang$^{1}$  \quad Jinwei Hu$^{1}$   \quad Xiangtai Li $^{2}$ \textsuperscript{$\dagger$}  \quad Yiwei He$^{1}$ \quad  \\ Xingyu Zhao$^{3}$ \quad Bei Peng$^{1}$ \quad Baoyuan Wu$^{4}$ \quad Xiaowei Huang$^{1}$ \quad Guangliang Cheng$^{1}$ \textsuperscript{$\dagger$}  \vspace{0.3em} \\
     {\normalsize $^1$University of Liverpool, UK \quad $^2$Nanyang Technological University, SG} \\
     {\normalsize $^3$ WMG, University of Warwick \quad $^4$The Chinese University of Hong Kong, Shenzhen,
Guangdong, China \quad} \\
    {\normalsize Project Page: \url{https://hzlsaber.github.io/projects/SIDA/}} \\
    {\normalsize \textsuperscript{$\dagger$} Corresponding author. E-mail: guangliang.cheng@liverpool.ac.uk \quad xiangtai94@gmail.com}
}

\begin{document}
\maketitle
\begin{abstract}
The rapid advancement of generative models in creating highly realistic images poses substantial risks for misinformation dissemination. For instance, a synthetic image, when shared on social media, can mislead extensive audiences and erode trust in digital content, resulting in severe repercussions. Despite some progress, academia has not yet created a large and diversified deepfake detection dataset for social media, nor has it devised an effective solution to address this issue. In this paper, we introduce the \textbf{S}ocial media \textbf{I}mage \textbf{D}etection data\textbf{Set} (SID-Set), which offers three key advantages: (1) \textbf{extensive volume}, featuring 300K AI-generated/tampered and authentic images with comprehensive annotations, (2) \textbf{broad diversity}, encompassing fully synthetic and tampered images across various classes, and (3) \textbf{elevated realism}, with images that are predominantly indistinguishable from genuine ones through mere visual inspection. Furthermore, leveraging the exceptional capabilities of large multimodal models, we propose a new image deepfake detection, localization, and explanation framework, named SIDA (\textbf{S}ocial media \textbf{I}mage \textbf{D}etection, localization, and explanation \textbf{A}ssistant). SIDA not only discerns the authenticity of images, but also delineates tampered regions through mask prediction and provides textual explanations of the model's judgment criteria. Compared with state-of-the-art deepfake detection models on SID-Set and other benchmarks, extensive experiments demonstrate that SIDA achieves superior performance among diversified settings. The code, model, and dataset will be released.
\end{abstract}    
\section{Introduction}
\label{sec:intro}
\par Recent advances in generative AI~\cite{DBLP:journals/pami/ZhanYWZLLKTX23, DBLP:journals/corr/abs-2406-14555, DBLP:journals/pami/CroitoruHIS23} have significantly improved the ability to generate highly realistic images, making it easier to create content that closely resembles real-world events. However, these advancements also bring new risks of malicious misuse, particularly in creating deceptive content aimed at misleading public opinion or distorting historical records. Such concerns have motivated the computer vision community to develop more sophisticated deepfake detection techniques. Contemporary methods~\cite{DBLP:conf/nips/00020YLW23, DBLP:journals/corr/abs-2402-00045} primarily focus on assessing the authenticity of facial images (i.e., {\it real} or {\it fake}), while an emerging subset aims to detect and localize facial manipulations~\cite{DBLP:conf/cvpr/GuoLRGM023, guo2024language}. These methods are typically trained on datasets containing real and fake images, aiming to detect images as real or fake, or to localize the tampered regions. Consequently, the quality and diversity of the datasets used for training and evaluation play a crucial role in achieving high accuracy in deepfake detection and localization. A well-curated dataset can enable models to learn nuanced features, improving robustness and generalization in real-world scenarios.

However, existing deepfake detection and localization datasets face two main challenges:
\begin{figure}[t]
  \centering
   \includegraphics[width=1.0\linewidth]{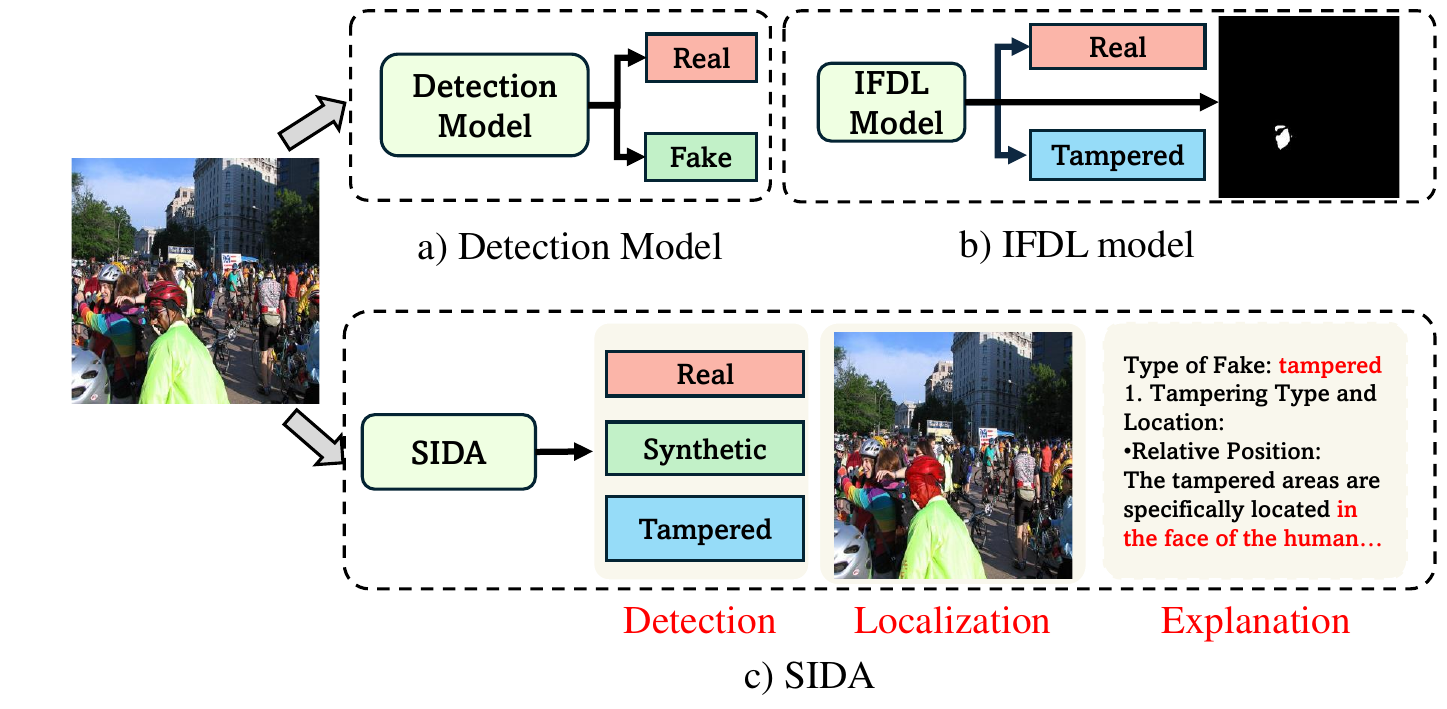}
   \caption{The framework comparisons. Existing deepfake methods (a-b) are limited to detection, localization, or both. In contrast, SIDA (c) offers a more comprehensive solution, capable of handling detection, localization, and explanation tasks.}
   \vspace{-1em}
   \label{sec1_figure1}
\end{figure}
\textbf{1) Insufficient Diversity.} The majority of existing datasets for deepfake detection focus mainly on facial imagery~\cite{DBLP:conf/mm/0002G0NK24, DBLP:conf/cvpr/HeGCZYSSS021}. However, given the growing capabilities of generative AI, the issue of non-facial image falsification on social media cannot be overlooked. While researchers~\cite{DBLP:conf/nips/ZhuCYHLLT0H023, DBLP:conf/iccv/WangBZWHCL23}  have developed relatively large datasets based on ImageNet for image deepfake detection, these datasets typically consist of images from simple scenarios that do not specifically focus on social media. Additionally, they often utilize somewhat outdated image-generation techniques, which can result in less convincing forgeries that are easier for both humans and models to detect.  Currently, there is a substantial lack of large-scale image deepfake datasets specifically designed for social media that leverage the latest generative methods.
\textbf{2) Limited Comprehensiveness.} 
Existing datasets are typically suited either for deepfake detection or for tampered region localization~\cite{10607909, DBLP:conf/cvpr/YuNLD024, DBLP:conf/cvpr/GuoLRGM023, guo2024language}, focusing on specific types of generative methods or image manipulations. However, an ideal deepfake dataset should encompass a wide range of scenarios to reflect the complexity of real social media content, where fake images may be fully generated or manipulated through image editing strategies~\cite{DBLP:journals/pami/ZhanYWZLLKTX23}. Furthermore, most existing datasets primarily focus on binary real/fake classification or tampered region localization, with limited emphasis on explaining the cues that models use to make these decisions. 

\begin{figure}[!ht]
\begin{center}
\centerline{\includegraphics[width=1.0\columnwidth]{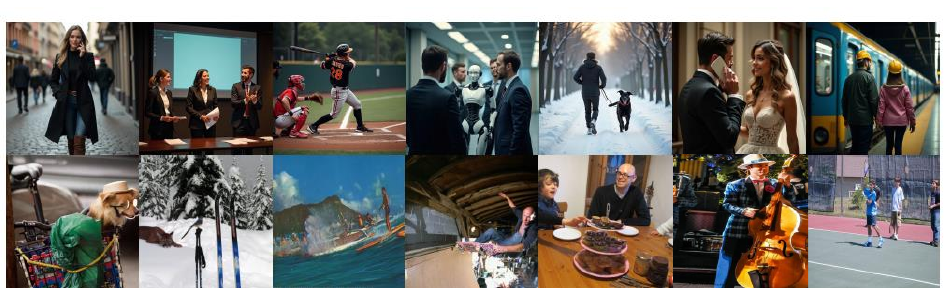}}
\caption{SID-Set examples. The 1st row is the synthetic images, while the 2nd row shows tampered images. (Zoom in to view)}
\label{sec2_Figure2}
\vspace{-1.5em} 
\end{center}
\end{figure}

To address these challenges, we introduce the \textbf{S}ocial media \textbf{I}mage \textbf{D}etection data\textbf{Set} (\textbf{SID-Set}), which consists of 300K images (\ie, 100K real, 100K synthetic, and 100K tampered images), providing a comprehensive resource for the deepfake detection community. Additionally, we include textual descriptions explaining the model's judgment basis. As shown in Figures~\ref{sec2_Figure2} and \ref{sec3_Figure4}, synthetic and tampered images are indistinguishable to humans. In particular, \textbf{challenges} for the SID-Set include: 1) subtle alterations of just dozens of pixels; 2) natural-looking local manipulations; 3) complex scenes in datasets. To our knowledge, SID-Set is the first dataset of its scale with extensive annotations, making it the largest and most comprehensive dataset for social media deepfake detection to date. Compared to existing datasets in Table~\ref{sec2:table1}, SID-Set addresses the challenges of limited diversity and outdated generative techniques by providing a more comprehensive set of high-quality and diverse images. Accordingly, we propose a new VLMs-based deepfake detection framework, named the \textbf{S}ocial media \textbf{I}mage \textbf{D}etection, localization, and explanation \textbf{A}ssistant (\textbf{SIDA}), which achieves the state-of-the-art (SOTA) performance on SID-Set and generalizes effectively across other benchmarks. SIDA can serve as a baseline model on SID-Set, offering a new framework for tackling social media image deepfake detection and localization.

The main contributions of this paper are as follows:
\begin{itemize}[leftmargin=0.5cm, itemindent=0cm]
\item We establish SID-Set, a comprehensive benchmark for detecting, localizing, and explaining deepfakes in social media images, featuring multiple image types and extensive annotations. SID-Set holds the potential for advancing the field of deepfake detection and ensuring robust model performance in complex real-world scenarios.
\item We propose SIDA, a new image deepfake detection, localization, and explanation framework that not only detects images with high accuracy but also localizes and explains potential manipulations, enhancing the transparency and utility of deepfake detection technologies.
\item Extensive experiments demonstrate that SIDA effectively identifies and delineates tampered areas within images, supporting the development of more robust and interpretable deepfake detection systems. Notably, SIDA demonstrates superior or equivalent performance on the SID-Set and other benchmarks.
\end{itemize}

\section{Related Work}
\label{sec:realted works}

\subsection{Image Deepfake Datasets}

In the realm of deepfake detection, the primary focus has historically centered on the identification of facial deepfakes. Renowned datasets such as ForgeryNet~\cite{DBLP:conf/cvpr/HeGCZYSSS021}, DeepFakeFace~\cite{DBLP:journals/corr/abs-2309-02218}, and DFFD~\cite{DBLP:conf/mm/0002G0NK24} have been pivotal in this area. As the field evolves, there is a growing shift among researchers towards exploring non-facial deepfake data. Advanced methodologies involving text-to-image or image-to-image generation techniques~\cite{DBLP:journals/corr/abs-2303-07909}, utilizing GANs or the stable diffusion series, have facilitated the creation of expansive deepfake datasets like GenImage~\cite{DBLP:conf/nips/ZhuCYHLLT0H023}, HiFi-IFDL~\cite{DBLP:conf/cvpr/GuoLRGM023}, and DiffForensics~\cite{DBLP:conf/iccv/WangBZWHCL23}. These datasets are characterized by their enlarged data volumes, diversified generation methodologies, and enriched annotation details. Furthermore, beyond the conventional real/fake annotations, certain datasets~\cite{DBLP:conf/cvpr/GuoLRGM023, DBLP:conf/cvpr/YuNLD024, guo2024language} now include more granular annotations. Table~\ref{sec2:table1} delineates a detailed comparison among various deepfake datasets, highlighting key differences in generation scenarios and annotation types supported. It shows that SID-Set is particularly tailored towards social media data, incorporating the latest SOTA generation models, emphasizing high-quality production, and providing much more comprehensive and diverse annotations.
\begin{table}[!t]
\caption{Comparison with existing image deepfake datasets.}
\label{sec2:table1}
\resizebox{0.47\textwidth}{!}{
\begin{tabular}{@{}ccccccc@{}}
\toprule
Dataset     & Content  & Data Source  & Generator Year & Multiclasses & Masks & Explanation \\ \midrule
OHImg~\cite{DBLP:conf/ih/MayTFS23}       & Overhead & Google Map   & 2023           & \textcolor{red}{\ding{55}}             & \textcolor{red}{\ding{55}}      &  \textcolor{red}{\ding{55}}           \\
FakeSpotter~\cite{DBLP:conf/ijcai/WangJMXHWL20} & Face     & CelebA, FFHQ        & 2020           & \textcolor{red}{\ding{55}}             & \textcolor{green}{\ding{51}}      &    \textcolor{red}{\ding{55}}         \\
ForgeryNet~\cite{DBLP:conf/cvpr/HeGCZYSSS021}  & Face     & CREMA-D       & 2021           &  \textcolor{red}{\ding{55}}             &   \textcolor{green}{\ding{51}}    &    \textcolor{red}{\ding{55}}          \\
DCFace~\cite{DBLP:conf/cvpr/Kim00023}      & Face     & FFHQ         & 2023           &          \textcolor{red}{\ding{55}}    &   \textcolor{red}{\ding{55}}    &      \textcolor{red}{\ding{55}}       \\
DFF~\cite{DBLP:journals/corr/abs-2309-02218}         & Face     & IMDB-WIKI     & 2023           &    \textcolor{red}{\ding{55}}            &  \textcolor{red}{\ding{55}}       & \textcolor{red}{\ding{55}}             \\
RealFaces~\cite{DBLP:conf/iwbf/PapaFCMA23}   & Face     & Prompts       & 2023           & \textcolor{red}{\ding{55}}              &   \textcolor{red}{\ding{55}}     &    \textcolor{red}{\ding{55}}          \\
M3Dsynth~\cite{DBLP:conf/icassp/ZingariniCCPV24}    & Biology  & DDPM, CycleGAN          & 2023           & \textcolor{red}{\ding{55}}              &  \textcolor{green}{\ding{51}}      &    \textcolor{red}{\ding{55}}          \\
CNNSpot~\cite{wang2019cnngenerated}     & Object   & SDXL          & 2020           & \textcolor{green}{\ding{51}}              &     \textcolor{red}{\ding{55}}   &   \textcolor{red}{\ding{55}}           \\
CiFAKE~\cite{DBLP:journals/access/BirdL24}     & Object   & CIFAR      & 2023           &    \textcolor{green}{\ding{51}}           &     \textcolor{red}{\ding{55}}    &  \textcolor{red}{\ding{55}}             \\
CASIA 2.0~\cite{Dong2013}   & General  & Corel         & 2022           & \textcolor{red}{\ding{55}}              &  \textcolor{green}{\ding{51}}      &    \textcolor{red}{\ding{55}}          \\
ArtiFact~\cite{DBLP:conf/icip/RahmanPSHF23}    & General  & COCO, FFHQ, LSUN & 2023        & \textcolor{red}{\ding{55}}              & \textcolor{red}{\ding{55}}       & \textcolor{red}{\ding{55}}             \\
IMD2020~\cite{DBLP:conf/wacv/NovozamskyMS20}     & General  & Places2       & 2020           &  \textcolor{red}{\ding{55}}            & \textcolor{green}{\ding{51}}      &  \textcolor{red}{\ding{55}}           \\
AIGCD~\cite{DBLP:journals/corr/abs-2311-12397}       & General  & LSUN, COCO, FFHQ & 2023        &  \textcolor{red}{\ding{55}}              &  \textcolor{red}{\ding{55}}       &   \textcolor{red}{\ding{55}}            \\
GenImage~\cite{DBLP:conf/nips/ZhuCYHLLT0H023}    & General  & ImageNet      & 2023           & \textcolor{red}{\ding{55}}               & \textcolor{red}{\ding{55}}        &  \textcolor{red}{\ding{55}}             \\
\textbf{SID-Set} & General  & COCO, Flickr30k, MagicBrush & 2024     & \textcolor{green}{\ding{51}}              &\textcolor{green}{\ding{51}}        & \textcolor{green}{\ding{51}}           \\ \bottomrule
\end{tabular}
}
\end{table}

\subsection{Image Deepfake Detection and Localization}
Deepfake detection methods~\cite{DBLP:journals/corr/abs-2403-17881, DBLP:journals/apin/MasoodNMJIM23, DBLP:conf/cvpr/WangD21, DBLP:conf/cvpr/ChenZSLW22, DBLP:conf/cvpr/Cao0YCDY22} are typically approached as classification tasks within the data-driven paradigm. These strategies primarily leverage diverse architectures~\cite{DBLP:journals/corr/abs-2403-17881, DBLP:journals/apin/MasoodNMJIM23}, including Convolutional Neural Networks (CNNs) and Transformers, to detect distinctive artifacts. Some scholars have attempted to achieve relatively high precision and generality by employing strategies such as employing
techniques such as data augmentation~\cite{DBLP:conf/cvpr/WangD21}, adversarial training~\cite{DBLP:conf/cvpr/ChenZSLW22},
reconstruction~\cite{DBLP:conf/cvpr/Cao0YCDY22}, etc. On the other hand, some researchers~\cite{DBLP:conf/aaai/JeongKRC22, DBLP:conf/aaai/Tan0WGLW24} have explored extracting features from the frequency domain for deepfake identification. Efforts~\cite{duan2024test, DBLP:conf/cvpr/WangY0HP23} have also been made to fuse features from both spatial and frequency domains to obtain a more comprehensive set of discriminative features for deepfake detection. Although these methods have shown some progress, they still struggle with issues of generalization. Furthermore, some scholars~\cite{DBLP:journals/corr/abs-2305-10794, DBLP:conf/cvpr/NguyenMSKA0GA24, DBLP:conf/cvpr/GuillaroCSDV23, DBLP:conf/mm/ZhangMLLDYLHFG024, DBLP:conf/wacv/TantaruOO24, DBLP:conf/iccv/ZhangXBZL0ALSS23} have gone beyond the basic classification between real and fake by gradually constructing datasets annotated with masks of locally tampered areas, thereby addressing both image deepfake detection and localization tasks. However, these datasets are concentrated mainly on facial data, with fewer datasets available for non-facial deepfake detection and localization, and even fewer large and public datasets for realistic social media data. As a result, our work aims to address these critical gaps by providing a new, comprehensive dataset that includes diverse manipulations beyond facial data, particularly focusing on social media images.
\subsection{Large Multimodal Models}
The progress in large language models (LLMs)~\cite{DBLP:journals/corr/abs-2402-06196, DBLP:journals/corr/abs-2302-13971, DBLP:journals/corr/abs-2307-09288, DBLP:journals/corr/abs-2407-21783} and vision-language models (VLMs)~\cite{DBLP:journals/pami/ZhangHJL24, DBLP:journals/corr/abs-2302-13971, DBLP:journals/corr/abs-2307-09288, DBLP:journals/corr/abs-2407-21783, DBLP:conf/cvpr/LaiTCLY0J24, DBLP:journals/corr/abs-2312-17240} has notably improved multimodal comprehension, seamlessly integrating visual and textual data. The LLaMA series~\cite{DBLP:journals/corr/abs-2302-13971, DBLP:journals/corr/abs-2307-09288, DBLP:journals/corr/abs-2407-21783} optimize language understanding with a compact, high-performance design using fewer parameters than prior models. LLaVA series~\cite{DBLP:conf/nips/LiuLWL23a, DBLP:conf/cvpr/LiuLLL24} enhance visual question answering by synchronizing visual features with textual data. Additionally, LISA series~\cite{DBLP:conf/cvpr/LaiTCLY0J24, DBLP:journals/corr/abs-2312-17240, li2023transformer, wu2024towards} employ LLM for accurate image segmentation, merging visual perception with linguistic insights to precisely segment the targeted areas. Several grounding large multimodal models~\cite{DBLP:conf/eccv/ZhangLLRZLHGLLY24, DBLP:journals/corr/abs-2305-18279, DBLP:conf/iclr/YouZGDZWCCY24, DBLP:conf/cvpr/XiaHHPSH24, DBLP:journals/corr/abs-2404-08506, DBLP:conf/cvpr/RenHW0FFJ24, DBLP:conf/cvpr/Rasheed0MS0CAX024, DBLP:journals/corr/abs-2306-14824, OMGLLaVA} have been proposed to localize the contents based on linguistic information.

Advancements~\cite{DBLP:journals/corr/abs-2310-17419, DBLP:conf/nips/Dai0LTZW0FH23, DBLP:conf/aaai/GuZZ00W24,DBLP:journals/corr/abs-2402-00126, DBLP:journals/corr/abs-2410-02761} in integrating multimodal data such as visual and linguistic information have also significantly improved the performance of models in detecting and pinpointing deepfakes. For instance, AntifakePrompt~\cite{DBLP:journals/corr/abs-2310-17419} approaches deepfake detection by formulating it as a visual question answering problem, which adjusts soft prompts for InstructBLIP~\cite{DBLP:conf/nips/Dai0LTZW0FH23}, enabling it to determine whether a query image is real or fake. ForgeryGPT~\cite{li2024forgerygpt} enhances image forgery detection and localization by integrating advanced forensic knowledge with a mask-aware forgery extractor that targets pixel-level fraud detection. As a concurrent work, FakeShield~\cite{DBLP:journals/corr/abs-2410-02761} leverages the capabilities of LLaVA to identify and localize altered regions while providing interpretable insights into the findings. Our method diverges significantly from existing approaches by creating the most extensive deepfake dataset tailored specifically for social media images with comprehensive annotations. Additionally, we also set a new standard for social media image deepfake detection by utilizing the visual interpretation strengths of VLMs to boost detection accuracy, pinpoint forgeries, and provide clearer explanations within a comprehensive framework.

\begin{figure*}[th]
\begin{center}
\centerline{\includegraphics[width=2.0\columnwidth]{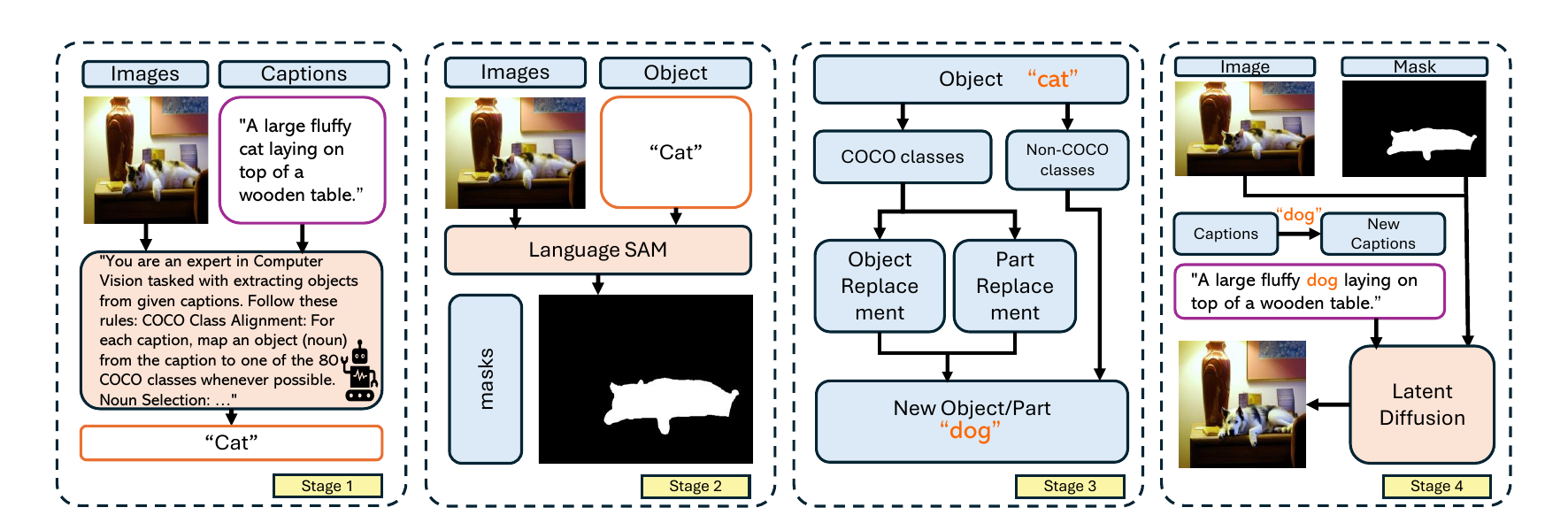}}
\caption{Tampered image generation pipeline: It consists of four stages—extracting objects from captions using GPT-4o, obtaining object masks with Language-SAM, setting up replacement dictionaries for generating tampered images, and generating new images using Latent Diffusion. This figure illustrates an example of object replacement (e.g., ``cat" to ``dog") and attribute modification.}
\label{sec3_Figure3}
\vspace{-2em} 
\end{center}
\end{figure*}

\section{Benchmark}
\subsection{Motivation}

In the realm of deepfake detection and localization, research has predominantly focused on facial deepfakes due to their substantial societal impact. However, with the advancements in generative technology, the scope of deepfakes has extended beyond facial content to include non-facial manipulations. Historically, non-facial deepfakes were less prevalent, largely limited by technological constraints that produced low-quality, easily detectable forgeries. Although some datasets, such as GenImage~\cite{DBLP:conf/nips/ZhuCYHLLT0H023} and AIGCD~\cite{DBLP:journals/corr/abs-2311-12397}, have been constructed, they suffer from several limitations: \textbf{1)} They often utilize relatively outdated generative technologies, resulting in lower quality data easily distinguished by humans. \textbf{2)} They primarily focus on text-to-image or image-to-image generation, neglecting the need for data involving manipulations of specific regions, objects, or parts. Such tampered manipulations can be especially insidious as they introduce subtle misinformation, making existing SOTA deepfake detection methods less effective. \textbf{3)} They lack well-defined criteria for content authenticity, limiting their effectiveness in providing interpretative insights and educating the public on distinguishing synthetic content.

Addressing these limitations is crucial for improving the transparency and utility of deepfake detection systems. Furthermore, most existing datasets emphasize either fully synthetic or tampered images, as shown in Table~\ref{sec2:table1}, but in real applications, we don't know the image deepfake type in advance. Effective deepfake detection and localization methods should be capable of addressing both scenarios, as social media images often involve complex combinations of synthetic and tampered content. Therefore, developing a comprehensive benchmark for detecting and localizing deepfakes in social media images is essential. We propose SID-Set, which encompasses comprehensive, high-quality annotations for detection and localization, along with detailed textual explanations of the judgment criteria.

\subsection{Benchmark Construction}
\label{sec:benchmark construction}
\textbf{Data Details.}  
To develop an effective benchmark for detecting and localizing images on social media, we created SID-Set, a dataset with real, synthetic, and tampered images reflecting diverse real-world scenarios. Our benchmark assesses whether models can differentiate among real, synthetic, and tampered images, as well as accurately identify altered regions in tampered images.

\textbf{Real Images:} 100K images from OpenImages V7\footnote{\url{https://storage.googleapis.com/openimages/web/index.html}}, with a wide range of scenarios reflecting real-world diversity.

\textbf{Synthetic Images:} 100K images generated through FLUX~\cite{DBLP:conf/iclr/Flux},  specifically designed to challenge identification due to their high-quality, highly realistic appearance.

\begin{figure}[!t]
\begin{center}
\centerline{\includegraphics[width=1.0\columnwidth]{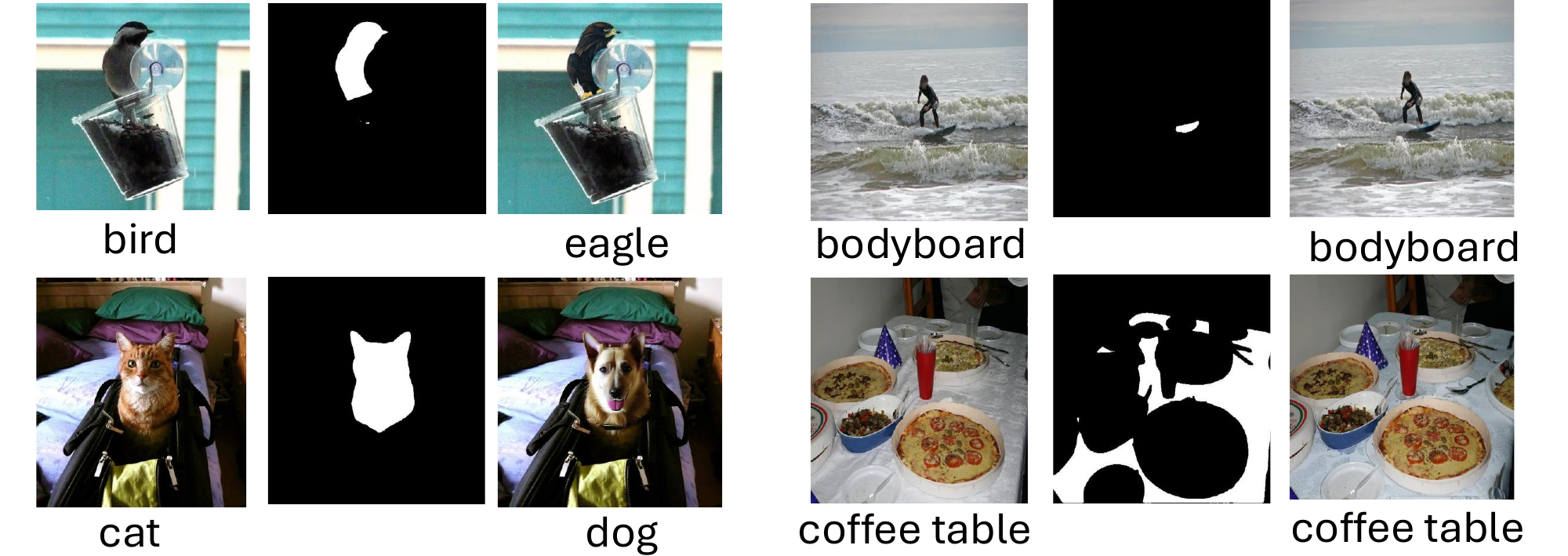}}
\caption{Examples of tampered images. (Zoom in to view)}
\label{sec3_Figure4}
\vspace{-3em} 
\end{center}
\end{figure}

\textbf{Tampered Images:} 100K tampered images, with specific objects or regions replaced or altered; the detailed generation process is shown in Figure~\ref{sec3_Figure4}.
\\
\textbf{Data Generation.}  
To generate highly realistic synthetic images, we experimented with several open-source SOTA generative models, such as FLUX~\cite{DBLP:conf/iclr/Flux}, Kandinsky 3.0~\cite{DBLP:journals/corr/abs-2312-03511}, SDXL~\cite{DBLP:conf/iclr/PodellELBDMPR24}, AbsoluteReality~\cite{DBLP:conf/iclr/Absolutereality}, and others. Following a review by five human experts of 1,000 images from each generative model, FLUX emerged as the top performer, producing highly convincing images that were indistinguishable from real ones to human experts. Consequently, we employed FLUX to create 100K synthetic images based on the Flickr30k~\cite{DBLP:journals/ijcv/PlummerWCCHL17} and COCO~\cite{DBLP:conf/eccv/LinMBHPRDZ14}. The image tampering process depicted in Figure~\ref{sec3_Figure4} follows four distinct stages, utilizing the COCO image as an example.

\textbf{Stage 1}: We extract objects from an image's caption using GPT-4o~\cite{DBLP:journals/corr/abs-2303-08774}. For instance, from the caption ``A large fluffy cat laying on top of a wooden table'', GPT-4o identifies relevant COCO class objects or retains nouns if no match exists. This extraction is documented in an ``Image-Caption-Object'' JSON file.

\textbf{Stage 2}: Employing Language-SAM~\cite{DBLP:conf/iclr/lang-sam}, we generate masks for identified objects as training ground truth.

\textbf{Stage 3}: We establish dictionaries for full and partial image tampering using COCO classes for object replacement and attribute modifications, respectively. For example, replacing "dog" with animals like "cat" or adding attributes such as "happy" or "angry" to the "dog" class. For more details, please refer to the Appendix.

\textbf{Stage 4}: Utilizing Latent Diffusion~\cite{DBLP:conf/cvpr/RombachBLEO22}, we modify captions and regenerate images, either replacing or retaining original objects based on availability. An example modification is altering "cat" to "dog" in the image caption.

In total, we generated 80,000 object-tampered images and 20,000 partially tampered images. To demonstrate the explainability of SIDA, we used GPT-4o to generate textual descriptions for the judgment basis of 3,000 images from the SID-Set, divided equally among three types. Additionally, to ensure the realism of synthetic images, tampered images, and their textual descriptions, we engaged 5 annotation experts for quality control and adjustments. Detailed prompts and generation pipelines used to create descriptions for each type of image are provided in the Appendix.

\section{Method}
\label{sec:Method}
\begin{figure*}[!t]
\begin{center}
\centerline{\includegraphics[width=1.8\columnwidth]{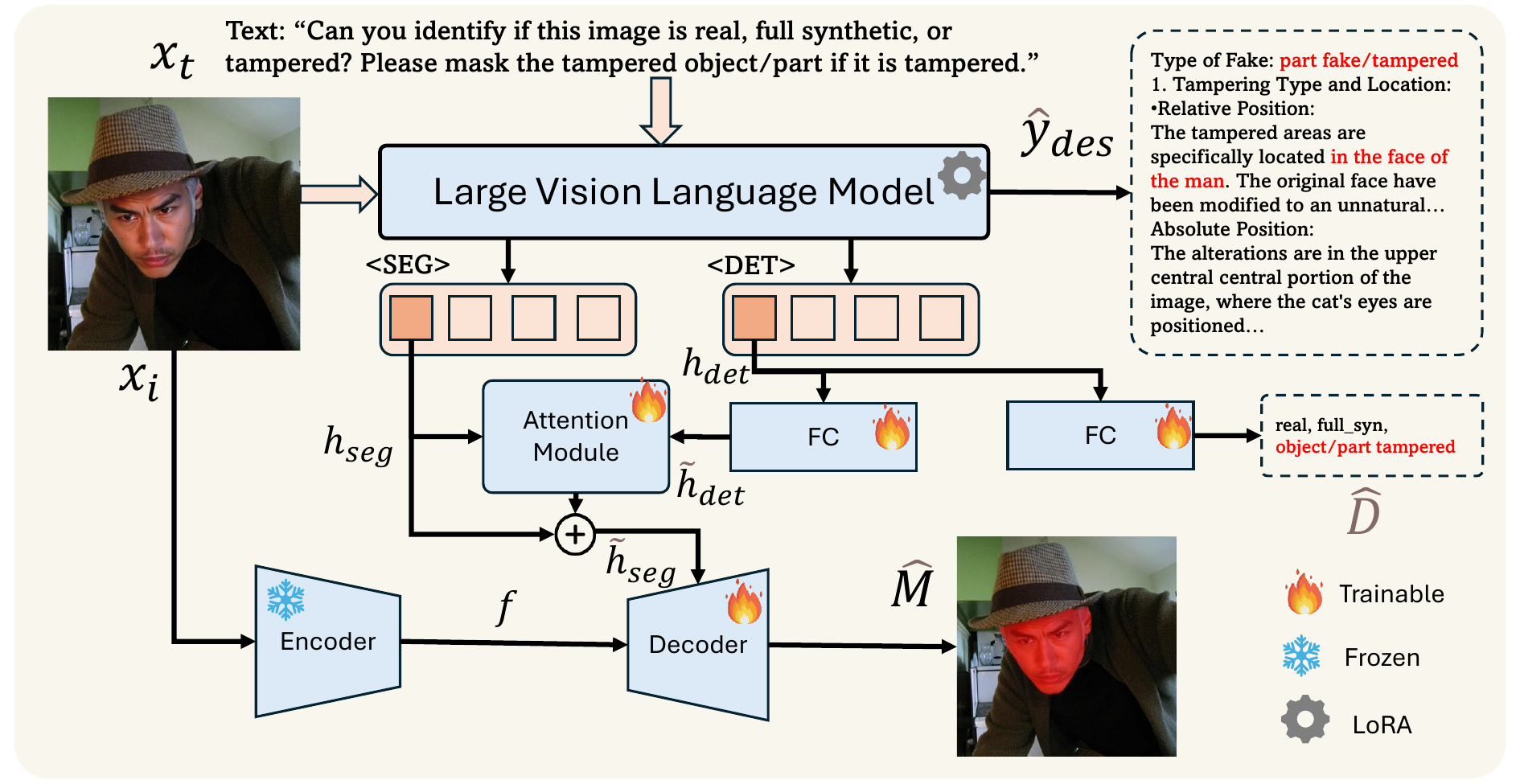}}
\caption{The pipeline of SIDA: Given an image $x_{i}$ and the corresponding text input $x_{t}$, the last hidden layer for the \texttt{<DET>} token provides the detection result. If the detection result indicates a tampered image, SIDA extracts the \texttt{<SEG>} token to generate masks for the tampered regions. This figure shows an example where the man's face has been manipulated.}
\vspace{-2em}
\label{sec3_Figure5}
\end{center}
\end{figure*}

In this section, we first present the model architecture of SIDA in Section \ref{sec:Architecture}, followed by an introduction to the training process of our method in Section \ref{sec:Training}.

\subsection{Architecture}
\label{sec:Architecture}
Large vision-language models have demonstrated remarkable capabilities in understanding the alignment between textual and visual information. For instance, LLaVA~\cite{DBLP:conf/nips/LiuLWL23a} leverages language alone to achieve a comprehensive understanding of both visual and linguistic content. Building on LLaVA, LISA~\cite{DBLP:conf/cvpr/LaiTCLY0J24} extends this capability by providing fine-grained segmentation masks along with corresponding textual descriptions. However, to effectively detect and localize synthetic images, VLMs must not only be capable of multimodal understanding but also possess the ability to identify and segment manipulated regions, providing detailed explanations for both synthetic and tampered images. 

To this end, we propose SIDA to tackle the task of synthetic image detection and tampered region localization. The pipeline of our method is illustrated in Figure~\ref{sec3_Figure5}. Inspired by previous approaches~\cite{DBLP:conf/cvpr/LaiTCLY0J24, DBLP:conf/icml/KimSK21, DBLP:conf/aaai/ZhouPZHCG20}, we expand the original vocabulary of VLMs by adding two new tokens, \texttt{<DET>} and \texttt{<SEG>}, to enable the model to extract detection and segmentation information. Given an image $x_{i}$ and a text prompt $x_{t}$, such as ``Can you identify if this image is real, fully synthetic, or tampered? Please mask the tampered object/part if it is tampered." We feed them into the VLM. The VLM then outputs a text description $\hat{y}_{\text{des}}$, while the last hidden layer $h_{\text{hid}}$ contain the \texttt{<DET>} and \texttt{<SEG>} tokens. This process can be formulated as follows:

\begin{equation}
    \hat{y}_{\text{des}} = \text{VLM}(x_{i}, x_{t}).
    \label{eq:input}
\end{equation}
Next, we extract the \texttt{<DET>} token from the last hidden layer $h_{\text{hid}}$ to obtain $h_{\text{det}}$. The representation $h_{\text{det}}$ is then passed through a detection head $F_{\texttt{det}}$ to determine whether the image is real, fully synthetic, or tampered. We denote the final detection result by $\hat{\texttt{D}}$:

\begin{equation} 
\hat{\texttt{D}} = F_{\texttt{det}}(h_{\text{det}}),
\label{eq:det_head} 
\end{equation}

\noindent $F_{\texttt{det}}$ is the detection head that processes the extracted \texttt{<DET>} representation $h_{\text{det}}$ to produce the detection output.

If the detection result indicates that the image has been tampered with, SIDA will then predict a mask for the tampered regions. The $h_{\text{seg}}$ feature is extracted from the hidden layer $h_{\text{hid}}$, similar to the extraction process for the \texttt{<DET>} token. Given that the \texttt{<DET>} token encapsulates crucial information that can aid in generating fine-grained segmentation masks, the representation $h_{\text{det}}$ is transformed using a fully connected layer $F$ to align with the dimensions of $h_{\text{seg}}$. To further capture the relationship between the $h_{\text{det}}$ and $h_{\text{seg}}$ features, we apply a single-layer Multihead Attention, facilitating effective feature interaction and enhancing mask quality.
The process can be formulated as follows:
\begin{equation}
    \begin{aligned}
        \tilde{h}_{\text{det}} &= F(h_{\text{det}}), \\
        \tilde{h}_{\text{seg}} &= \texttt{MSA}(\tilde{h}_{\text{det}}, h_{\text{seg}}), \\
        \tilde{h}_{\text{seg}} &= \tilde{h}_{\text{seg}} + h_{\text{seg}}.
    \end{aligned}
    \label{eq:attention}
\end{equation}
In this formulation, we treat the detection features as the query, and the segmentation features as the key and value. A residual connection is employed to combine both the original and transformed information, thereby enhancing the representation for precise segmentation.

Finally, we employ a frozen image encoder $F_{\texttt{enc}}$ to extract visual features from the input image $x_{i}$, resulting in visual features $f$. The segmentation embedding $\tilde{h}_{\text{seg}}$ is then combined with the visual features $f$ and fed into a decoder to produce the final mask $\hat{M}$. This can be formulated as:

\begin{equation}
    \begin{aligned}
    f &= F_{\texttt{enc}}(x_{i}), \\
    \hat{\texttt{M}} &= F_{\texttt{dec}}(\tilde{h}_{\text{seg}}, f).
    \end{aligned}
\end{equation}

\subsection{Training}
\label{sec:Training}
\textbf{Training Objectives. }
The training loss, $\mathcal{L}$, for the SIDA consists of three components: the detection loss $\mathcal{L}_{det}$, the text generation loss $\mathcal{L}_{txt}$, and the segmentation mask loss $\mathcal{L}_{mask}$. Initially, SIDA is trained in an end-to-end manner by employing the detection loss and the segmentation loss. For detection, we use CrossEntropy loss, while for the segmentation task, we use a weighted combination of binary cross-entropy (BCE) and DICE loss, with respective loss weights $\lambda{\text{bce}}$ and $\lambda_{\text{dice}}$. This can be formulated as:
\begin{equation}
\begin{aligned}
\mathcal{L} &= \lambda_{det} \mathcal{L}_{det} + \lambda_{mask} \mathcal{L}_{mask}, \\
\mathcal{L}_{det} &= \mathcal{L}_{\text{CE}}(\hat{\texttt{D}} ,\texttt{D} ), \\
\mathcal{L}_{mask} &= \lambda_{bce} \mathcal{L}_{BCE}(\hat{\texttt{M}},\texttt{M}) + \lambda_{dice}\mathcal{L}_{DICE}(\hat{\texttt{M}},\texttt{M}).
\label{eq:loss_1} 
\end{aligned}
\end{equation}

After completing the training phase, we proceed to fine-tune the SIDA model by utilizing detailed textural descriptions from 3,000 images as the ground truth, represented by $y_{des}$. This phase focuses on optimizing the text generation component, $\mathcal{L}_{txt}$, to improve its ability in textural interpretability. The final loss function is as follows:
\begin{equation}
\begin{aligned}
\mathcal{L}_{txt} &= \mathcal{L}_{\text{CE}}(\hat{y_{des}},y_{des} ), \\
\mathcal{L}_{total} &= \lambda_{det} \mathcal{L}_{det} + \lambda_{mask} \mathcal{L}_{mask} + \lambda_{txt} \mathcal{L}_{txt} ,
\label{eq:total_loss} 
\end{aligned}
\end{equation}
where $\lambda_{det}$, $\lambda_{mask}$, and $\lambda_{txt}$ are the weighting factors that balance the contributions of the detection, segmentation, and text generation losses, respectively.
\\
\textbf{Training Data.}
We utilize the SID-Set, consisting of 300k images, to train SIDA. To further enhance diversity, we incorporate the MagicBrush dataset~\cite{DBLP:conf/nips/ZhangMCSS23} after filtering out low-quality images. The combined dataset supports robust training for both the detection and localization of synthetic content. Additionally, we generate descriptions for 3,000 randomly selected images using LLMs.

\section{Experiments}
\label{sec:Experiments}

\textbf{Implementation Details. }
We choose LISA as the base large vision language model due to its strong capability for reasoning-based localization. We fine-tuned both LISA-7B-v1 and LISA-13B-v1 on the SID-Set using LoRA, setting $\alpha$ at 16 and the dropout rate at 0.05. The input images are resized to $1024 \times 1024$. The loss weights in Eq.~(\ref{eq:total_loss}) for the detection ($\lambda_{\text{det}}$), the text generation ($\lambda_{\text{txt}}$), and the localization ($\lambda_{\text{mask}}$) are set to 1.0, respectively. The localization loss weights in Eq.~(\ref{eq:loss_1}) for the $\lambda_{\text{bce}}$ and $\lambda_{\text{dice}}$ are set to 1.0. To determine the optimal weight configuration, we perform ablation studies as detailed in Section \ref{subsec:Ablation Study}.
During the detection and localization training stage, the image encoder is frozen, and all other modules are trainable. For the text generation stage, only vision-language models are fine-tuned using the LoRA strategy. The initial learning rate is set to $1 \times 10^{-4}$, with a batch size of 2 per device and a gradient accumulation step of 10. We use two NVIDIA A100 GPUs (40GB each). Training for SIDA-7B and SIDA-13B took 48 hours and 72 hours, respectively. We divided the SID-Set into a training set, validation set, and test set using a 7:1:2 ratio.

\begin{table*}[ht]
    \caption{Comparison of SIDA with other deepfake detection methods. Values outside parentheses are evaluated with open-source models directly on the SID-Set; values in parentheses indicate performance changes after fine-tuning the models using the SID-Set on the validation set. ${\textcolor{red}{Red}}$ indicates improvement, and ${\textcolor{green}{green}}$ indicates a decrease in performance. The best results are in bold. The overall accuracy and F1 are calculated as the average of the values from the three classification categories.}
    \small
    \setlength{\tabcolsep}{4pt} 
    \centering
    \begin{tabular}{ccccccccccc}
        \Xhline{1pt}
         \multirow{2}{*}{Methods}&\multirow{2}{*}{Year} &\multicolumn{2}{c}{Real} & \multicolumn{2}{c}{Fully synthetic} & \multicolumn{2}{c}{Tampered}& \multicolumn{2}{c}{Overall} \\
         \cmidrule(r){3-4}
         \cmidrule(r){5-6}
         \cmidrule(r){7-8}
         \cmidrule(r){9-10}
         & &Acc & F1 &Acc & F1&Acc & F1&Acc & F1 \\
        \midrule
        AntifakePrompt~\cite{DBLP:journals/corr/abs-2310-17419} &2024  
        & 64.8${\textcolor{red}{\scriptstyle(\uparrow24.1)}}$ 
        & 78.6${\textcolor{red}{\scriptstyle(\uparrow10.5)}}$  
        & 93.8${\textcolor{red}{\scriptstyle(\uparrow3.7)}}$ \phantom 
        & 96.8${\textcolor{red}{\scriptstyle(\uparrow1.1)}}$  \phantom
        & 30.8${\textcolor{red}{\scriptstyle(\uparrow60.1)}}$  
        & 47.2${\textcolor{red}{\scriptstyle(\uparrow33.2)}}$  
        & 63.1${\textcolor{red}{\scriptstyle(\uparrow29.3)}}$  
        & 74.2${\textcolor{red}{\scriptstyle(\uparrow14.9)}}$
        \\
        CnnSpott~\cite{DBLP:journals/corr/abs-2104-02984} 
        &2021 
        & 79.8${\textcolor{red}{\scriptstyle(\uparrow9.2)}}$ \phantom
        & 88.7${\textcolor{red}{\scriptstyle(\uparrow2.1)}}$ \phantom 
        & 39.5${\textcolor{red}{\scriptstyle(\uparrow51.2)}}$  
        & 56.6${\textcolor{red}{\scriptstyle(\uparrow31.5)}}$ 
        & 6.9${\textcolor{red}{\scriptstyle(\uparrow61.2)}}$  \phantom
        & 12.9${\textcolor{red}{\scriptstyle(\uparrow51.1)}}$  
        & 42.1${\textcolor{red}{\scriptstyle(\uparrow40.5)}}$  
        & 52.7${\textcolor{red}{\scriptstyle(\uparrow28.2)}}$  
        \\
        FreDect~\cite{DBLP:conf/icml/FrankESFKH20} 
        &2020  
        & 83.7${\textcolor{green}{\scriptstyle(\downarrow37.7)}}$ 
        
        & 91.1${\textcolor{green}{\scriptstyle(\downarrow43.5)}}$ 
        & 16.8${\textcolor{red}{\scriptstyle(\uparrow44.1)}}$  
        & 28.8${\textcolor{red}{\scriptstyle(\uparrow37.2)}}$  
        & 11.9${\textcolor{red}{\scriptstyle(\uparrow25.2)}}$  
        & 21.3${\textcolor{red}{\scriptstyle(\uparrow31.7)}}$   
        & 37.4${\textcolor{red}{\scriptstyle(\uparrow10.2)}}$   
        & 47.0${\textcolor{red}{\scriptstyle(\uparrow8.4)}}$   
        \\
       Fusing~\cite{DBLP:conf/icip/JuJKXNL22} 
       &2022 
       & 85.1${\textcolor{red}{\scriptstyle(\uparrow4.1)}}$ \phantom 
       & \textbf{92.0${\textcolor{red}{\scriptstyle(\uparrow0.7)}}$ } \phantom
       & 34.0${\textcolor{red}{\scriptstyle(\uparrow54.1)}}$  
       & 50.7${\textcolor{red}{\scriptstyle(\uparrow38.4)}}$  
       & 2.7${\textcolor{red}{\scriptstyle(\uparrow24.3)}}$  
       & 5.3${\textcolor{red}{\scriptstyle(\uparrow26.1)}}$  
       & 40.6${\textcolor{red}{\scriptstyle(\uparrow27.5)}}$  
       & 49.3${\textcolor{red}{\scriptstyle(\uparrow21.7)}}$ 
       \\
        Gram-Net~\cite{DBLP:conf/cvpr/LiuQT20} 
        &2020 
        & 70.1${\textcolor{red}{\scriptstyle(\uparrow19.1)}}$ 
        & 82.4${\textcolor{red}{\scriptstyle(\uparrow9.3)}}$ \phantom 
        & 93.5${\textcolor{red}{\scriptstyle(\uparrow4.4)}}$ 
        \phantom
        & 96.6${\textcolor{red}{\scriptstyle(\uparrow2.0)}}$  
        \phantom
        & 0.8${\textcolor{red}{\scriptstyle(\uparrow89.1)}}$ \phantom 
        & 1.6${\textcolor{red}{\scriptstyle(\uparrow85.3)}}$ \phantom 
        & 54.8${\textcolor{red}{\scriptstyle(\uparrow37.3)}}$  
        & 60.2${\textcolor{red}{\scriptstyle(\uparrow32.2)}}$ 
        \\
        UnivFD~\cite{DBLP:conf/cvpr/OjhaLL23} 
        & 2023
        & 68.0${\textcolor{red}{\scriptstyle(\uparrow0.3)}}$ \phantom
        & 67.4${\textcolor{red}{\scriptstyle(\uparrow1.1)}}$ \phantom
        & 62.1${\textcolor{red}{\scriptstyle(\uparrow24.3)}}$ 
        & 87.5${\textcolor{red}{\scriptstyle(\uparrow10.5)}}$
        & 64.0${\textcolor{red}{\scriptstyle(\uparrow28.5)}}$ 
        & 85.3${\textcolor{red}{\scriptstyle(\uparrow4.7)}}$ \phantom
        & 64.7${\textcolor{red}{\scriptstyle(\uparrow17.7)}}$ 
        & 80.0${\textcolor{red}{\scriptstyle(\uparrow5.4)}}$
        \phantom{0} 
        \\
        LGrad~\cite{DBLP:conf/cvpr/Tan0WGW23} 
        & 2023 
        & 64.8${\textcolor{green}{\scriptstyle(\downarrow2.8)}}$ \phantom 
        & 78.6${\textcolor{green}{\scriptstyle(\downarrow2.5)}}$ \phantom
        & 83.5${\textcolor{green}{\scriptstyle(\downarrow25.5)}}$
        & 91.0${\textcolor{green}{\scriptstyle(\downarrow23.7)}}$
        & \textbf{6.8${\textcolor{red}{\scriptstyle(\uparrow\textbf{92.3})}}$} 
        & \textbf{12.7${\textcolor{red}{\scriptstyle(\uparrow\textbf{86.1})}}$}
        & 51.7${\textcolor{red}{\scriptstyle(\uparrow21.3)}}$
        & 60.7${\textcolor{red}{\scriptstyle(\uparrow19.9)}}$
        \\
        LNP~\cite{DBLP:journals/corr/abs-2311-00962} 
        & 2023
        & 71.2${\textcolor{green}{\scriptstyle(\downarrow56.8)}}$
        & 83.2${\textcolor{green}{\scriptstyle(\downarrow60.2)}}$ 
        & 91.8${\textcolor{green}{\scriptstyle(\downarrow55.6)}}$ 
        & 95.7${\textcolor{green}{\scriptstyle(\downarrow60.1)}}$ 
        & 2.9${\textcolor{red}{\scriptstyle(\uparrow90.4)}}$ \phantom
        & 5.7${\textcolor{red}{\scriptstyle(\uparrow88.9)}}$ \phantom
        & 55.3${\textcolor{green}{\scriptstyle(\downarrow7.3)}}$
        \phantom
        & 61.5${\textcolor{green}{\scriptstyle(\downarrow10.4)}}$ 
        \phantom{0} 
        \\
        \hline
        SIDA-7B 
        &2024 
        & 89.1
        &91.0
        & \textbf{98.7} 
        & 98.6
        & 92.7
        & 91.0 
        & 93.5 
        & \textbf{93.5}
        \\
        SIDA-13B 
        & 2024
        & \textbf{89.6}
        & 91.1
        & 98.5 
        & \textbf{98.7}
        & 92.9
        & 91.2 
        & \textbf{93.6}
        & \textbf{93.5}
        \\
        \Xhline{1pt}
        
    \end{tabular}
    \label{tabel:detection}
\end{table*}

\noindent
\textbf{Evaluation Metrics.}
We evaluate detection using image-level accuracy and F1 scores. For forgery localization, our metrics include Area Under the Curve (AUC), F1 scores, and Intersection over Union (IoU).
\subsection{Detection Evaluation}
We compare SIDA against other SOTA deepfake detection methods on SID-Set, including CnnSpot~\cite{DBLP:journals/corr/abs-2104-02984}, AntifakePrompt~\cite{DBLP:journals/corr/abs-2310-17419}, FreDect~\cite{DBLP:conf/icml/FrankESFKH20}, Fusing~\cite{DBLP:conf/icip/JuJKXNL22}, Gram-Net~\cite{DBLP:conf/cvpr/LiuQT20}, UnivFD~\cite{DBLP:conf/cvpr/OjhaLL23}, LGrad~\cite{DBLP:conf/cvpr/Tan0WGW23}, and LNP~\cite{DBLP:journals/corr/abs-2311-00962}. To ensure a fair comparison, we first evaluate these models on our dataset using their original pre-trained weights, then retrain them with the SID-Set to assess performance improvements. Table~\ref{tabel:detection} demonstrates that SIDA achieves better or comparable results among all the evaluated methods. Notably, LGrad~\cite{DBLP:conf/cvpr/Tan0WGW23} achieves the highest accuracy and F1 score on tampered images after retraining, but this comes at the expense of lower performance in other metrics. Our analysis indicates that LGrad's high recall and false positive rates stem from its propensity to misclassify other types as tampered. The training details are provided in the Appendix.
\label{subsec:Detection Results}
\subsection{Localization Results}
\label{subsec:Localization Results}
Table~\ref{tabel:localization} presents the forgery localization performance on the SID-Set. We selected PSCC-Net~\cite{DBLP:journals/tcsv/LiuLCL22}, MVSS-Net~\cite{DBLP:journals/pami/DongCHCL23}, and HIFI-Net~\cite{DBLP:conf/cvpr/GuoLRGM023} as representative IFDL methods. Additionally, we chose LISA~\cite{DBLP:conf/cvpr/LaiTCLY0J24} as a representative LLM due to its segmentation reasoning capabilities. We used LISA-7B-v1 and fine-tuned it on SID-Set. The results indicate that SIDA achieves the best performance. We suppose that while LISA possesses strong general segmentation capabilities, it lacks the specific specialized features required to detect subtle manipulations, ultimately limiting the effectiveness of fine-tuning for precise forgery localization.
\begin{table}[!ht]
    \vspace{-0.5em}
    \caption{Comparison between SIDA and other IFDL approaches. * indicates the use of the pre-trained model from the original paper due to unavailable training code.}
    \renewcommand{\arraystretch}{1.1}
    \centering
    \begin{tabular}{cccll}
\Xhline{1pt}
\multirow{2}{*}{Methods} & \multirow{2}{*}{Years} & \multicolumn{3}{c}{\multirow{2}{*}{Tampered}} \\
                         &                        & \multicolumn{3}{c}{}                          \\ \cline{3-5} 
\multicolumn{1}{l}{}     & \multicolumn{1}{l}{}   & AUC            & F1           & IOU           \\ \hline
MVSS-Net*~\cite{DBLP:journals/pami/DongCHCL23}                & 2023                   &          48.9      &     31.6         &    23.7           \\
HIFI-Net*~\cite{DBLP:conf/cvpr/GuoLRGM023}                & 2023                   &  64.0              &       45.9       &    21.1           \\
PSCC-Net~\cite{DBLP:journals/tcsv/LiuLCL22}                 & 2022                   &         82.1       &      71.3        &   35.7            \\
LISA-7B-v1~\cite{DBLP:conf/cvpr/LaiTCLY0J24}     & 2024                   & 78.4              &  69.1          &  32.5            \\
SIDA-7B                     & 2024                   &   \textbf{87.3}              &        \textbf{73.9}     &        \textbf{43.8}     \\ \Xhline{1pt}
    \end{tabular}
    \label{tabel:localization}
    \vspace{-1em}
\end{table}
\subsection{Robustness Study}
\label{subsec:Robustness Test}
We further evaluate the robustness of SIDA against common image perturbations found in social media, such as JPEG compression, resizing, and Gaussian noise. Table \ref{table:robust} shows our model's performance on the SID-Set under six degradation scenarios: JPEG compression (with quality levels of 70 and 80), resizing (with scaling factors of 0.5 and 0.75), and Gaussian noise (with variances of 5 and 10). Despite not being explicitly trained on degraded data, SIDA demonstrates resilience to these low-level distortions. The model's stable performance against common social media perturbations highlights its robustness and practical applicability.
\begin{table}[!ht]
    \caption{Performance of SIDA under different perturbations.}
    \vspace{-0.5em}
    \renewcommand{\arraystretch}{1.0}
    \centering
    \begin{tabular}{cccccc}
    \Xhline{1pt}
    \multirow{2}{*}{} & \multicolumn{2}{c}{Detection} & \multicolumn{3}{c}{Localization} \\ \cline{2-6} 
                      & ACC            & F1           & AUC       & F1      & IOU       \\ \hline
    JPEG 70           &  89.4              &   90.1           &  86.2      &   71.8    &   42.3     \\
    JPEG 80           &  88.7              &   89.5           &  85.8      &   71.1    &   41.7     \\
    Resize  0.5       &  89.3              &   91.1           &  86.8      &   72.5    &   43.2     \\
    Resize  0.75      &  89.9              &   91.6           &  87.1      &   73.0    &   43.5     \\
    Gaussian 10       &  86.9              &   89.3           &  84.1      &   70.2    &   41.0     \\
    Gaussian 5        &  88.4              &   89.9           &  85.3      &   71.0    &   41.5     \\
    SIDA-7B           & 93.5               & 93.5             &  87.3      &   73.9    &   43.8     
  \\ \Xhline{1pt}
        \end{tabular}
    \vspace{-0.5em}
    \label{table:robust}
\end{table}

\subsection{Test on Other Benchmark}
\label{subsec:Benchmark}
In this stage, we evaluate SIDA on DMimage~\cite{DBLP:conf/icassp/CorviCZPNV23} dataset to assess its generalization capabilities. We compare SIDA with CNNSpot~\cite{DBLP:journals/corr/abs-2104-02984}, Fusing~\cite{DBLP:conf/icip/JuJKXNL22}, Gram-Net~\cite{DBLP:conf/cvpr/LiuQT20}, LNP~\cite{DBLP:journals/corr/abs-2311-00962}, UnivFD~\cite{DBLP:conf/cvpr/OjhaLL23}, and AntifakePrompt~\cite{DBLP:journals/corr/abs-2310-17419}. For these methods, we use the original hyperparameter settings and pre-trained weights provided by the authors. The results in Table~\ref{table:benchmark} show that SIDA achieves superior performance, demonstrating its strong adaptability.

\begin{table}[!ht]
\centering
\caption{Comparison with other deepfake detection methods on DMimage~\cite{DBLP:conf/icassp/CorviCZPNV23}. We evaluate each method using its original pre-trained weights and use an equal number of real and fake images sampled from DMImage.}
\label{table:benchmark}
\scalebox{0.85}{ 
\begin{tabular}{ccccccc}
\Xhline{1pt}
\multirow{2}{*}{Methods} & \multicolumn{2}{c}{Real} & \multicolumn{2}{c}{Fake} & \multicolumn{2}{c}{Overall} \\ 
\cmidrule(r){2-3} \cmidrule(r){4-5} \cmidrule(r){6-7}
 & Acc & F1 & Acc & F1 & Acc & F1 \\ \midrule
CNNSpot~\cite{DBLP:journals/corr/abs-2104-02984}   &  87.8   & 88.4   &  28.4  & 44.2   & 58.1 & 66.3   \\
Gram-Net~\cite{DBLP:conf/cvpr/LiuQT20}  &  62.8   & 54.1   &  78.8  & 88.1   & 70.8 & 71.1   \\
Fusing~\cite{DBLP:conf/icip/JuJKXNL22}   &  87.7   & 86.1   &  15.5  & 27.2   & 51.6 & 56.6   \\
LNP~\cite{DBLP:journals/corr/abs-2311-00962}  &  63.1   & 67.4   &  56.9  & 72.5   & 60.0 & 70.0   \\
UnivFD~\cite{DBLP:conf/cvpr/OjhaLL23}  &  89.4   & 88.3   &  44.9  & 61.2   & 67.2 & 74.8   \\
AntifakePrompt~\cite{DBLP:journals/corr/abs-2310-17419} &  91.3   & 92.5   &  89.3  & \textbf{91.2}   & 90.3 & 91.8   \\
SIDA-7B &  \textbf{92.9}   & \textbf{93.1}   &  \textbf{90.7}  & 91.0   & \textbf{91.8} & \textbf{92.0}   \\
\Xhline{1pt}

\end{tabular}
}
\end{table}
\vspace{-1em}

\subsection{Ablation Study}
\label{subsec:Ablation Study}

\textbf{Attention Module. }
We conducted ablation experiments to assess the importance of the attention module. Variants included removing the attention module entirely and replacing it with fully connected (FC) layers. Results in Table~\ref{table:ablation_attention} show that removing the attention module or replacing it with FC layers significantly reduces performance, underscoring the critical role of attention in enhancing feature interaction and improving detection and localization accuracy.
\begin{table}[!ht]
    \centering
    \vspace{-0.5em} 
    \caption{Ablation study results for attention module in SIDA.}
    \begin{tabular}{cccccc}
    \Xhline{1pt}
    \multirow{2}{*}{} & \multicolumn{2}{c}{Detection} & \multicolumn{3}{c}{Localization} \\ \cline{2-6} 
                      & ACC            & F1           & AUC       & F1      & IOU       \\ \hline
        FC & 91.1 & 90.3 & 84.3 & 71.6 & 38.9 \\
        w/o Attention & 90.3 & 89.9 & 84.1 & 71.3 & 38.8 \\
        SIDA & 93.5               & 93.5             &  87.3      &   73.9    &   43.8      \\
        \bottomrule
    \end{tabular}
    \label{table:ablation_attention}
    
\end{table}
\\
\textbf{Training Weights.} SIDA training utilizes weighted losses to balance task contributions. In the detection and localization stages, detection loss is adjusted by a weight $\lambda_{\text{det}}$, and localization loss by binary cross-entropy (BCE) and DICE losses, with weights $\lambda_{\text{bce}}$ and $\lambda_{\text{dice}}$ respectively. For our experiments, we set $\lambda_{\text{det}}$ to 1, $\lambda_{\text{bce}}$ to 2.0, and $\lambda_{\text{dice}}$ to 0.5 to maintain balance between detection and localization, enhancing model stability and performance. We summarize the outcomes of various weight configurations in Table~\ref{table:ablation_weights}.
\begin{table}[!ht]
    \centering
    \caption{Ablation detection results for different weight configurations in SIDA training.}
    \begin{tabular}{cccccc}
        \toprule
        \textbf{$\lambda_{\text{det}}$} & \textbf{$\lambda_{\text{bce}}$} & \textbf{$\lambda_{\text{dice}}$} & Acc & F1 Score \\
        \midrule
        1.0 & 2.0 & 0.5 & 93.56 &  91.01\\
        1.0 & 4.0 & 1.0 & 93.49& 90.86 \\
        \bottomrule
    \end{tabular}
    \label{table:ablation_weights}
    \vspace{-1em} 
\end{table}

\subsection{Qualitative Results}
\label{subsec:Qualitative Results}
\begin{figure}[!t]
\begin{center}
\centerline{\includegraphics[width=1.08\columnwidth]{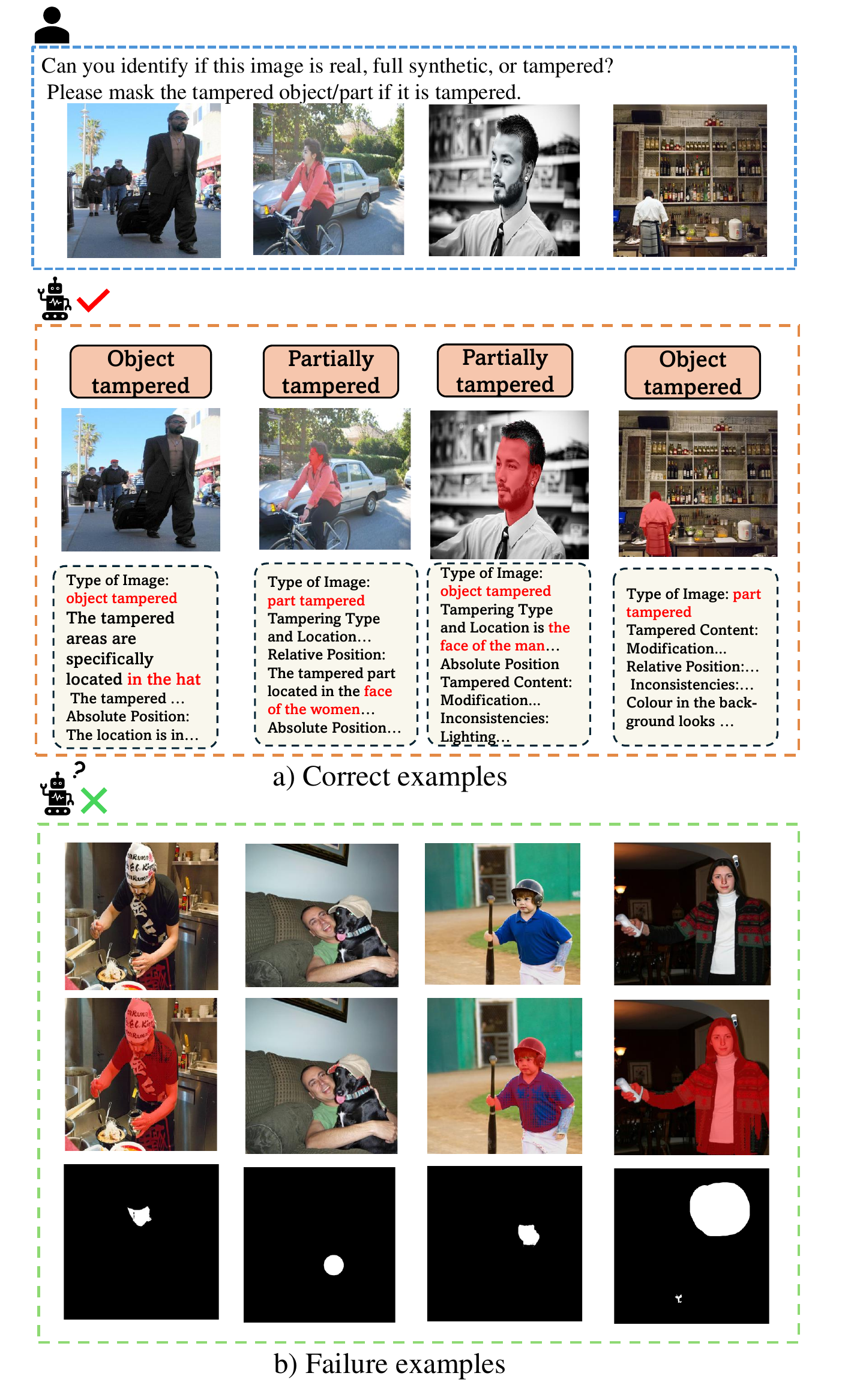}}
\caption{Visual results of SIDA on tampered images.}
\label{sec5_Figure6}
\vspace{-3.5em} 
\end{center}
\end{figure}

In this section, we present examples of SIDA's output for tampered images, showcasing its detection, localization, and explanation capabilities. SIDA accurately identifies tampered regions and provides explanations for its decisions. We also include some challenging failure cases where SIDA was unable to accurately detect the tampered regions, highlighting areas for future improvement. Additional visualization results are available in the Appendix.
\section{Conclusion and Discussions}
\label{sec:conclusion}
In this work, we present SID-Set for social image deepfake detection, localization, and explanation tasks, consisting of 100k real images, 100k fully synthetic images, and 100k tampered images. Furthermore, we propose a new VLMs-based deepfake detection framework, SIDA, to address these tasks. SIDA demonstrates its ability to detect fake types, localize tampered regions, and provide explanations for its decisions. We believe that the integration of VLMs into deepfake detection tasks, as demonstrated by SIDA, offers promising new avenues for future research in this critical field. 

Although the development of the SID-Set and the introduction of the SIDA framework have yielded favorable outcomes in deepfake detection tasks, we recognize some potential limitations and will optimize them in future research.
\\
\textbf{Dataset Size}.
While SID-Set includes 100k fully synthetic and 100k tampered images, the complexity of real social media environments demands a larger dataset. Therefore, expanding the dataset with additional images is a crucial objective for future research. 
\\
\textbf{Data Domain}.
Other methods often generate social media images that lack authenticity or are easily identifiable. Consequently, we used only FLUX to generate fake images due to its superior quality. However, depending exclusively on one method may lead to issues of data skew, although this was not significantly evident in our experiments. Such skew could potentially impair performance on diverse datasets. Moving forward, we plan to explore additional generative methods and integrate various other generation techniques to produce a more varied and higher-quality set of images.
\\
\textbf{Localization Results}.
Although SIDA demonstrates relatively strong performance on our dataset, there is still room for improvement. Certain tampered regions are not reliably detected, underscoring the need for further advancements.

\section{Appendix}
\label{suppl:appendix}
\appendix
\textbf{Contents of the Appendices:}

Section~\ref{suppl:exp_config}. Details of Experimental Settings, Hyperparameters, and Configurations.

Section~\ref{suppl:comparison}. Detailed Comparison with Related Work.

Section~\ref{suppl:visualization}.  Additional Visual Examples, Including Failures and Failure Analysis of SIDA.

Section~\ref{suppl:generation}. Detailed DataSet Creation Process.

Section~\ref{suppl:experts}. Experts and Human Evaluation.

\section{Experiment Settings}
\label{suppl:exp_config}

\textbf{Detection Methods.}
We used AIGCDetecBenchmark\footnote{\url{https://github.com/Ekko-zn/AIGCDetectBenchmark}} GitHub to test and re-train CnnSpot~\cite{DBLP:journals/corr/abs-2104-02984}, FreDect~\cite{DBLP:conf/icml/FrankESFKH20}, Fusing~\cite{DBLP:conf/icip/JuJKXNL22}, 
Gram-Net~\cite{DBLP:conf/cvpr/LiuQT20}, UnivFD~\cite{DBLP:conf/cvpr/OjhaLL23}, LGrad~\cite{DBLP:conf/cvpr/Tan0WGW23}, and LNP~\cite{DBLP:journals/corr/abs-2311-00962}. For AntifakePrompt~\cite{DBLP:journals/corr/abs-2310-17419},we used the original training settings provided in the official GitHub repository\footnote{\url{https://github.com/nctu-eva-lab/AntifakePrompt}}. During testing and training, we used only classification labels for these detection methods, as they cannot handle localization tasks.

We set noise (\eg JPEG compression, blur, and resize) to None for testing each approach. For CNNSpot~\cite{DBLP:journals/corr/abs-2104-02984}, FreDect~\cite{DBLP:conf/icml/FrankESFKH20}, Fusing~\cite{DBLP:conf/icip/JuJKXNL22}, and Gram-Net~\cite{DBLP:conf/cvpr/LiuQT20}, we retrained them with the following hyperparameters: a blur probability of 0.1 with a sigma range of 0.0 to 3.0, a JPEG compression probability of 0.1, and JPEG quality ranging from 30 to 100. We used a batch size of 64, a crop size of 224, and Adam as the optimizer. We used different hyperparameters to achieve the best results for LGrad~\cite{DBLP:conf/cvpr/Tan0WGW23}, LNP~\cite{DBLP:journals/corr/abs-2311-00962}, and UnivFD~\cite{DBLP:conf/cvpr/OjhaLL23}, which require image pre-processing. 
Specifically, for LNP and LGrad, both the blur probability and JPEG compression probability were set to 0. For UnivFD, we used the same training settings as CNNSpot after pre-processing. For AntifakePrompt, we used the same hyperparameters and prompts as described in the original paper, recording and calculating performance across different classes in the results. All methods were trained for 10 epochs on a single NVIDIA A100 40GB GPU. Methods that did not require image pre-processing took approximately 36 hours to train, while LGrad, LNP, and UnivFD, which needed pre-processing, took around 48 hours.
\\
\textbf{Localization Methods.}
We used the pre-trained models for MVSS-Net~\cite{DBLP:journals/pami/DongCHCL23} and HIFI-Net~\cite{DBLP:conf/cvpr/GuoLRGM023} to evaluate performance on SID-Set. For PSCC-Net~\cite{DBLP:journals/tcsv/LiuLCL22}, we used the same training settings as provided in the official GitHub repository\footnote{\url{https://github.com/proteus1991/PSCC-Net/tree/main}}. For LISA~\cite{DBLP:conf/cvpr/LaiTCLY0J24}, we used the LISA-7B-v1 version and fine-tuned it on SID-Set for comparison. Specifically, we set the learning rate to 0.0001, the batch size to 2, and the gradient accumulation steps to 10.

\begin{table}[!t]
\caption{Comparison with existing related works. An (*) indicates methods that have created their own dataset.}
\label{suppl:table8}
\resizebox{0.47\textwidth}{!}{
\begin{tabular}{ccccccc}
\toprule
\multirow{2}{*}{Methods} & \multirow{2}{*}{Year} & \multirow{2}{*}{Has dataset*} & \multicolumn{2}{c}{Detection} & \multirow{2}{*}{Localization} & \multirow{2}{*}{Interpretation} \\ \cline{4-5}
                         &                       &                              & Binary  & Muti-classification &                               &                                 \\ \midrule
DIRE~\cite{DBLP:conf/iccv/WangBZWHCL23}                 
& 2023                
&\textcolor{green}{\ding{51}}                  
&\textcolor{green}{\ding{51}}      
& \textcolor{red}{\ding{55}}                      
&\textcolor{red}{\ding{55}}         
&\textcolor{red}{\ding{55}}              \\
AntifakePrompt~\cite{DBLP:journals/corr/abs-2310-17419}       
& 2024                 
&\textcolor{red}{\ding{55}}                      
&\textcolor{green}{\ding{51}}         
& \textcolor{red}{\ding{55}}                      
&\textcolor{red}{\ding{55}}                         
&\textcolor{red}{\ding{55}}                        \\
CnnSpott~\cite{DBLP:journals/corr/abs-2104-02984}              
& 2021                 
&\textcolor{green}{\ding{51}}              
&\textcolor{green}{\ding{51}}    
&\textcolor{red}{\ding{55}}                       
&\textcolor{red}{\ding{55}}           
&\textcolor{red}{\ding{55}}           \\
FreDect~\cite{DBLP:conf/icml/FrankESFKH20}               
& 2020                 
& \textcolor{red}{\ding{55}}        
&\textcolor{green}{\ding{51}}
& \textcolor{red}{\ding{55}}                      
&\textcolor{red}{\ding{55}}             
&\textcolor{red}{\ding{55}}             \\
Fusing~\cite{DBLP:conf/icip/JuJKXNL22}                
& 2022               
& \textcolor{red}{\ding{55}}            
&\textcolor{green}{\ding{51}}    
& \textcolor{red}{\ding{55}}                      
&\textcolor{red}{\ding{55}}           
&\textcolor{red}{\ding{55}}           \\
Gram-Net~\cite{DBLP:conf/cvpr/LiuQT20}              
& 2020                 
& \textcolor{red}{\ding{55}}                      
& \textcolor{green}{\ding{51}}         
& \textcolor{red}{\ding{55}}                      
& \textcolor{red}{\ding{55}}                         
&\textcolor{red}{\ding{55}}                      \\
UnivFD~\cite{DBLP:conf/cvpr/OjhaLL23}                
& 2023                
&\textcolor{green}{\ding{51}}              
&\textcolor{green}{\ding{51}}    
& \textcolor{red}{\ding{55}}                      
&\textcolor{red}{\ding{55}}           
&\textcolor{red}{\ding{55}}           \\
LGrad~\cite{DBLP:conf/cvpr/Tan0WGW23}                
& 2023                 
&\textcolor{red}{\ding{55}}                      
&\textcolor{green}{\ding{51}}         
&\textcolor{red}{\ding{55}}                       
&\textcolor{red}{\ding{55}}                         
&\textcolor{red}{\ding{55}}                      \\
LNP~\cite{DBLP:journals/corr/abs-2311-00962}                   
& 2023               
& \textcolor{green}{\ding{51}}              
& \textcolor{green}{\ding{51}}    
&\textcolor{red}{\ding{55}}                       
& \textcolor{red}{\ding{55}}           
& \textcolor{red}{\ding{55}}           \\
MVSS-Net~\cite{DBLP:journals/pami/DongCHCL23}              
& 2023                
&\textcolor{red}{\ding{55}}                       
&\textcolor{green}{\ding{51}}         
&\textcolor{red}{\ding{55}}                       
&\textcolor{green}{\ding{51}}                      
&\textcolor{red}{\ding{55}}              \\
HIFI-Net~\cite{DBLP:conf/cvpr/GuoLRGM023}               
& 2023                 
&\textcolor{green}{\ding{51}}                   
&\textcolor{green}{\ding{51}}         
&\textcolor{red}{\ding{55}}                       
&\textcolor{green}{\ding{51}}                      
&\textcolor{red}{\ding{55}}       \\
PSCC-Net~\cite{DBLP:journals/tcsv/LiuLCL22}             
& 2022                 
&\textcolor{green}{\ding{51}}                   
&\textcolor{green}{\ding{51}}         
&\textcolor{red}{\ding{55}}                       
&\textcolor{green}{\ding{51}}                      
&\textcolor{red}{\ding{55}}          \\
FFAA~\cite{DBLP:journals/corr/abs-2408-10072}                 
& 2024                 
&\textcolor{green}{\ding{51}}                   
&\textcolor{green}{\ding{51}}         
&\textcolor{red}{\ding{55}}                       
&\textcolor{green}{\ding{51}}                      
&\textcolor{green}{\ding{51}}           \\
FakeShield~\cite{DBLP:journals/corr/abs-2410-02761}           
& 2024                 
&\textcolor{green}{\ding{51}}                   
&\textcolor{green}{\ding{51}}         
&\textcolor{red}{\ding{55}}                       
&\textcolor{green}{\ding{51}}                      
&\textcolor{green}{\ding{51}}           \\
ForgeryGPT~\cite{li2024forgerygpt}           
& 2024                 
&\textcolor{green}{\ding{51}}                   
&\textcolor{green}{\ding{51}}         
&\textcolor{red}{\ding{55}}                       
&\textcolor{green}{\ding{51}}                      
&\textcolor{green}{\ding{51}}           \\
SIDA                 
& 2024                 
&\textcolor{green}{\ding{51}}                   
&\textcolor{green}{\ding{51}}         
&\textcolor{green}{\ding{51}}                        
&\textcolor{green}{\ding{51}}                      
&\textcolor{green}{\ding{51}}               \\ \bottomrule
\end{tabular}
}
\end{table}

\section{Detailed Comparison}
\label{suppl:comparison}
Due to page limitations, we selected only a few representative works for the main comparison. In this section, we present a more comprehensive comparison of SIDA with additional related works, as shown in Table~\ref{suppl:table8}.

Compared to detection methods~\cite{DBLP:journals/corr/abs-2310-17419, DBLP:conf/iccv/WangBZWHCL23, DBLP:journals/corr/abs-2104-02984, DBLP:conf/icml/FrankESFKH20, DBLP:conf/icip/JuJKXNL22, DBLP:conf/cvpr/LiuQT20, DBLP:conf/cvpr/OjhaLL23, DBLP:conf/cvpr/Tan0WGW23, DBLP:journals/corr/abs-2311-00962}, which often specialize in identifying specific generative techniques, SIDA is designed with a broader focus, capable of handling various manipulation types. This versatility allows SIDA to generalize better across different datasets and manipulations, making it more effective in real-world scenarios. Additionally, SIDA provides both detection and localization, offering a more comprehensive solution compared to detection-only models.

Compared to existing IFDL (Image Forgery Detection and Localization) methods~\cite{DBLP:journals/pami/DongCHCL23, DBLP:conf/cvpr/GuoLRGM023, DBLP:journals/tcsv/LiuLCL22}, which primarily focus on detecting tampered versus real images, SIDA is capable of handling a broader range of scenarios, including fully synthetic, tampered, and real images. This allows SIDA to provide a more comprehensive detection capability. Furthermore, SIDA leverages LLMs to enhance the interpretability of its localization results, delivering not only segmentation masks but also detailed explanations. This combination improves precision and adds a valuable interpretative layer that existing methods lack, making it effective for understanding and addressing manipulations in complex scenarios.

Compared to other works that have explored the use of LLMs in deepfake detection, our approach addresses multiple tasks, utilizes a larger dataset, and produces fine-grained outputs.  For example, compared with FFAA~\cite{DBLP:journals/corr/abs-2408-10072}, SID-Set is not limited to facial deepfake detection but is designed to tackle more complex scenarios commonly found on social media, such as object manipulation and partial tampering. Compared with FakeShield~\cite{DBLP:journals/corr/abs-2410-02761}, our work not only includes more realistic images but also provides detailed fine-grained results, thereby enhancing detection accuracy and interpretability. Additionally, compared to ForgeryGPT~\cite{li2024forgerygpt}, we have curated a large-scale, high-quality dataset that serves as a valuable resource to support and advance research in this domain.

Since some works have not released their code until paper submission~\cite{DBLP:journals/corr/abs-2408-10072, DBLP:journals/corr/abs-2410-02761, li2024forgerygpt}, we chose PSCC-Net~\cite{DBLP:journals/tcsv/LiuLCL22} and LISA~\cite{DBLP:conf/cvpr/LaiTCLY0J24} to demonstrate localization results due to their effective segmentation capabilities. We retrained both models on SID-Set for 10 epochs and obtained the output results. Compared to these methods, SIDA shows superior performance in detecting the borders of tampered areas, delivering more precise and clearer results, as illustrated in Figure~\ref{suppl:comp}.
\begin{figure*}[!t]
\begin{center}
\centerline{\includegraphics[width=2.0\columnwidth]{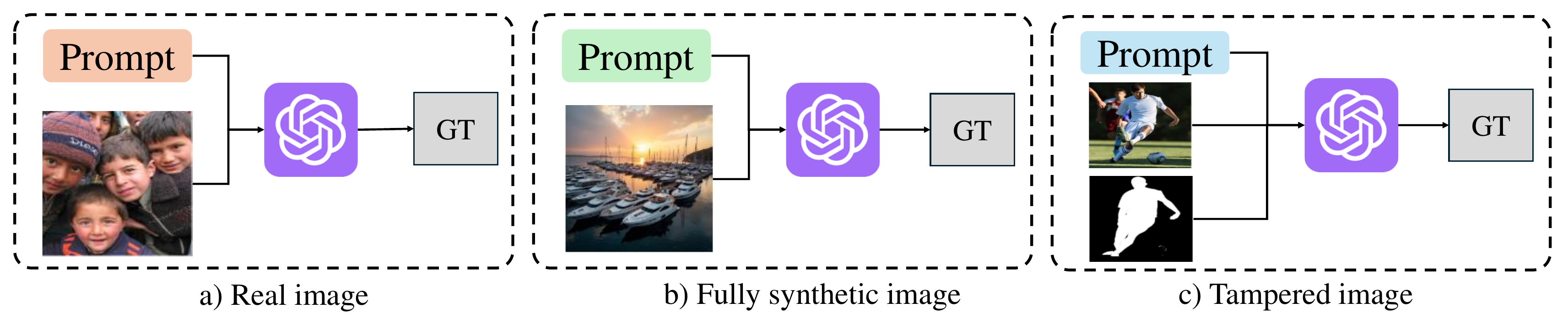}}
\caption{Generating ground truth descriptions for three different types of inputs.}
\label{suppl:Figure18} 
\end{center}
\vspace{-3em}
\end{figure*}

\begin{figure*}[!ht]
\begin{center}
\centerline{\includegraphics[width=1.7\columnwidth]{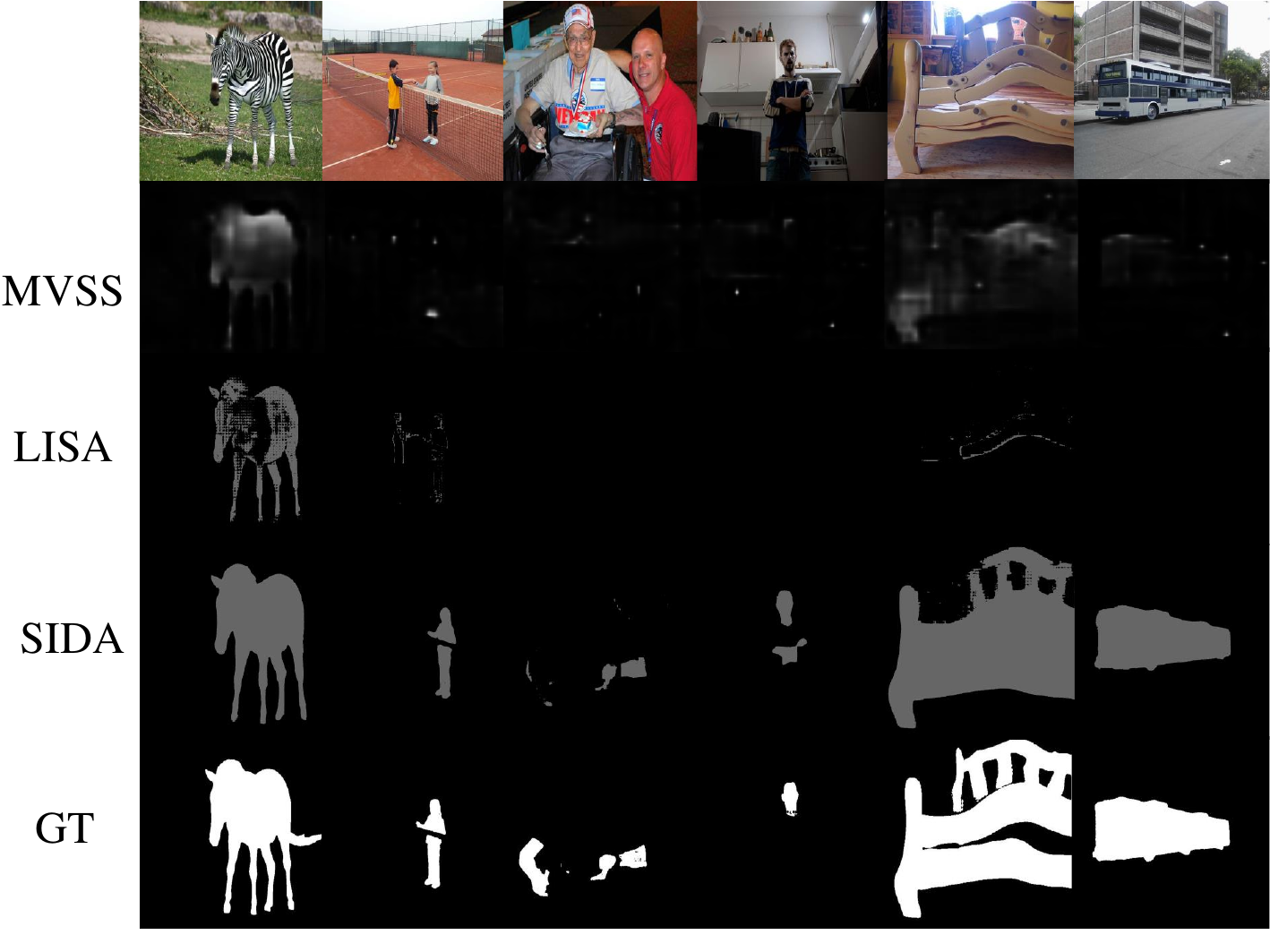}}
\caption{Visual comparison of SIDA with other localization methods. Both approaches were fine-tuned on the SID-Set for this evaluation.}
\label{suppl:comp} 
\vspace{-3em}
\end{center}
\end{figure*}

\section{Additional Visual Examples}
\label{suppl:visualization}

\begin{figure*}[b]
\begin{center}
\centerline{\includegraphics[width=2\columnwidth]{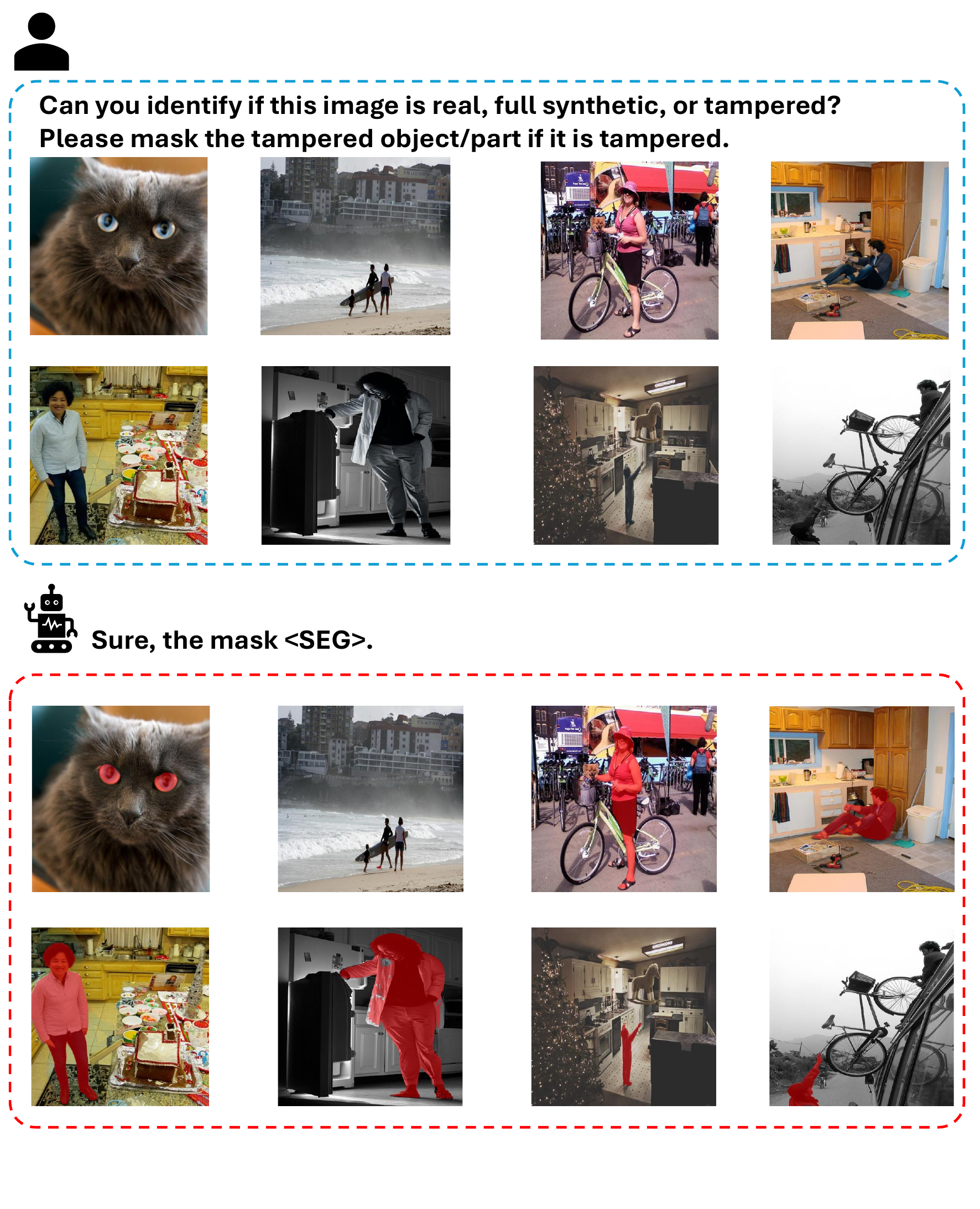}}
\caption{Visual examples of SIDA.}
\label{suppl:Figure15} 
\end{center}
\end{figure*}

\begin{figure*}[b]
\begin{center}
\centerline{\includegraphics[width=2\columnwidth]{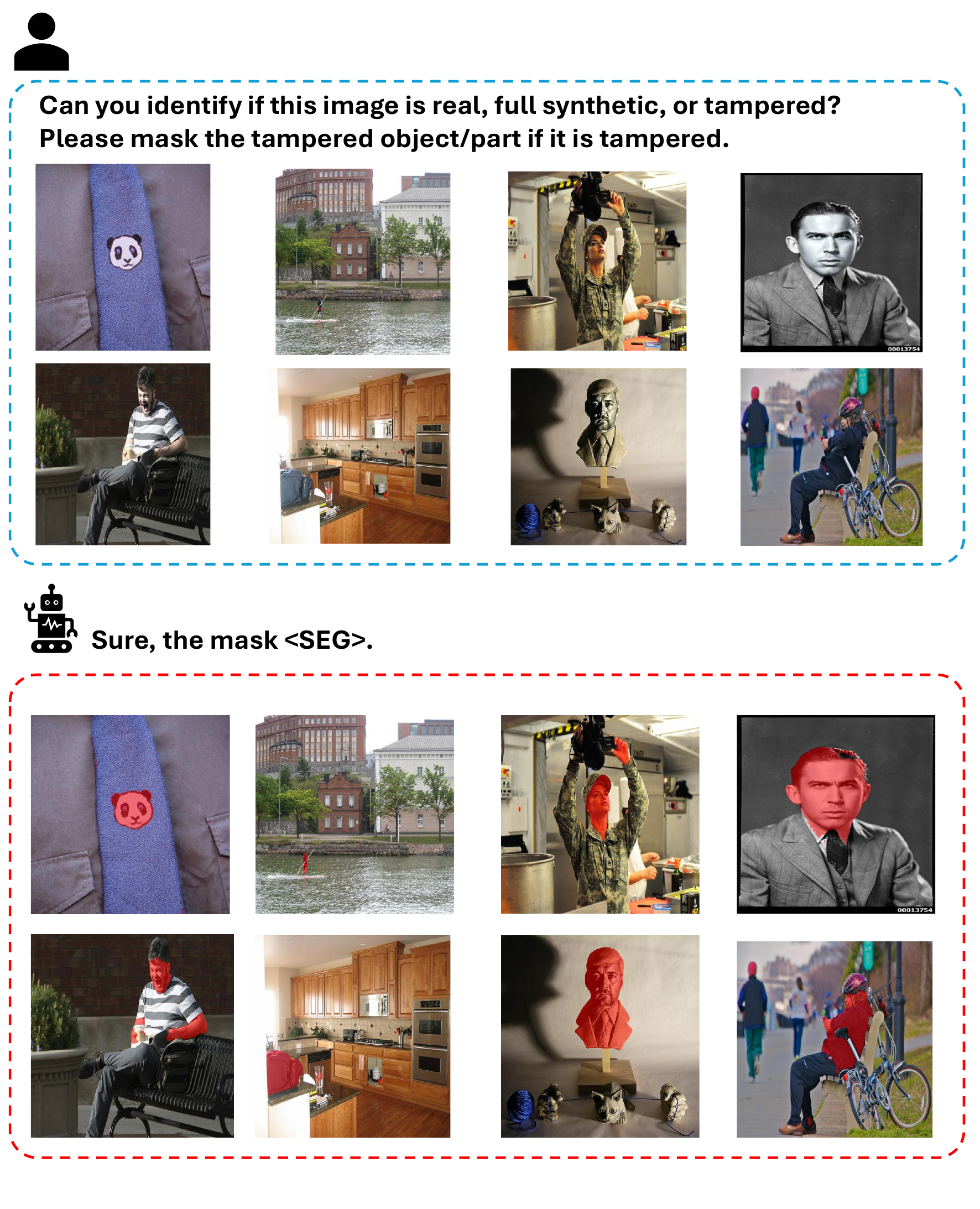}}
\caption{More Visual examples of SIDA.}
\label{suppl:Figure16} 
\end{center}
\end{figure*}

\begin{figure*}[b]
\begin{center}
\centerline{\includegraphics[width=1.45\columnwidth]{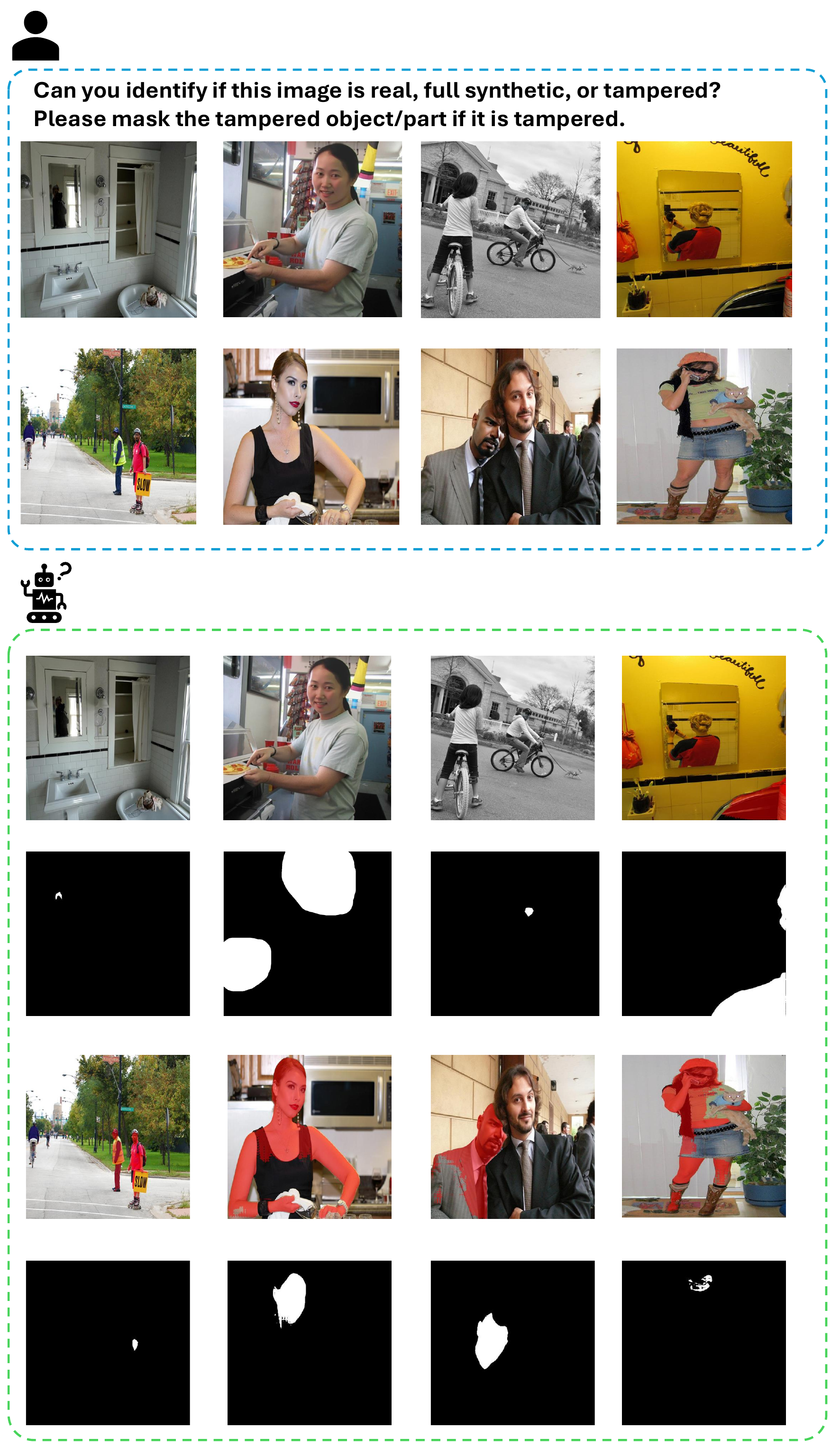}}
\caption{Failure cases of SIDA.}
\label{suppl:Figure17} 
\end{center}
\end{figure*}

In this section, we provide additional visual examples of SIDA. Figures~\ref{suppl:Figure15} and~\ref{suppl:Figure16} depict SIDA's outputs for tampered images, while Figure~\ref{suppl:Figure17} highlights some failure cases. 

The first row in Figure~\ref{suppl:Figure17} illustrates instances where SIDA fails to detect tampered areas, with some cases resulting in no mask output at all. The second row demonstrates SIDA's inability to generate fine-grained masks for the tampered regions. We attribute these shortcomings to two primary factors. First, the current training data for tampered images may be insufficient. Although SID-Set provides 100k tampered images, this volume might still be inadequate for the LLM to effectively handle highly detailed and complex manipulations. Second, although SIDA surpasses other methods in detecting tampered regions, it may still lack the precision required for particularly challenging cases involving subtle or intricate tampering. These limitations indicate critical areas for future research. We aim to improve both the quality and quantity of training data, while also developing more sophisticated methodologies and enhancement strategies to better address the challenges posed by complex manipulation scenarios, ultimately enhancing detection accuracy and mask quality.

\section{Detailed DataSet Creation Process}
\label{suppl:generation}
\begin{figure*}[b]
\begin{center}
\centerline{\includegraphics[width=2\columnwidth]{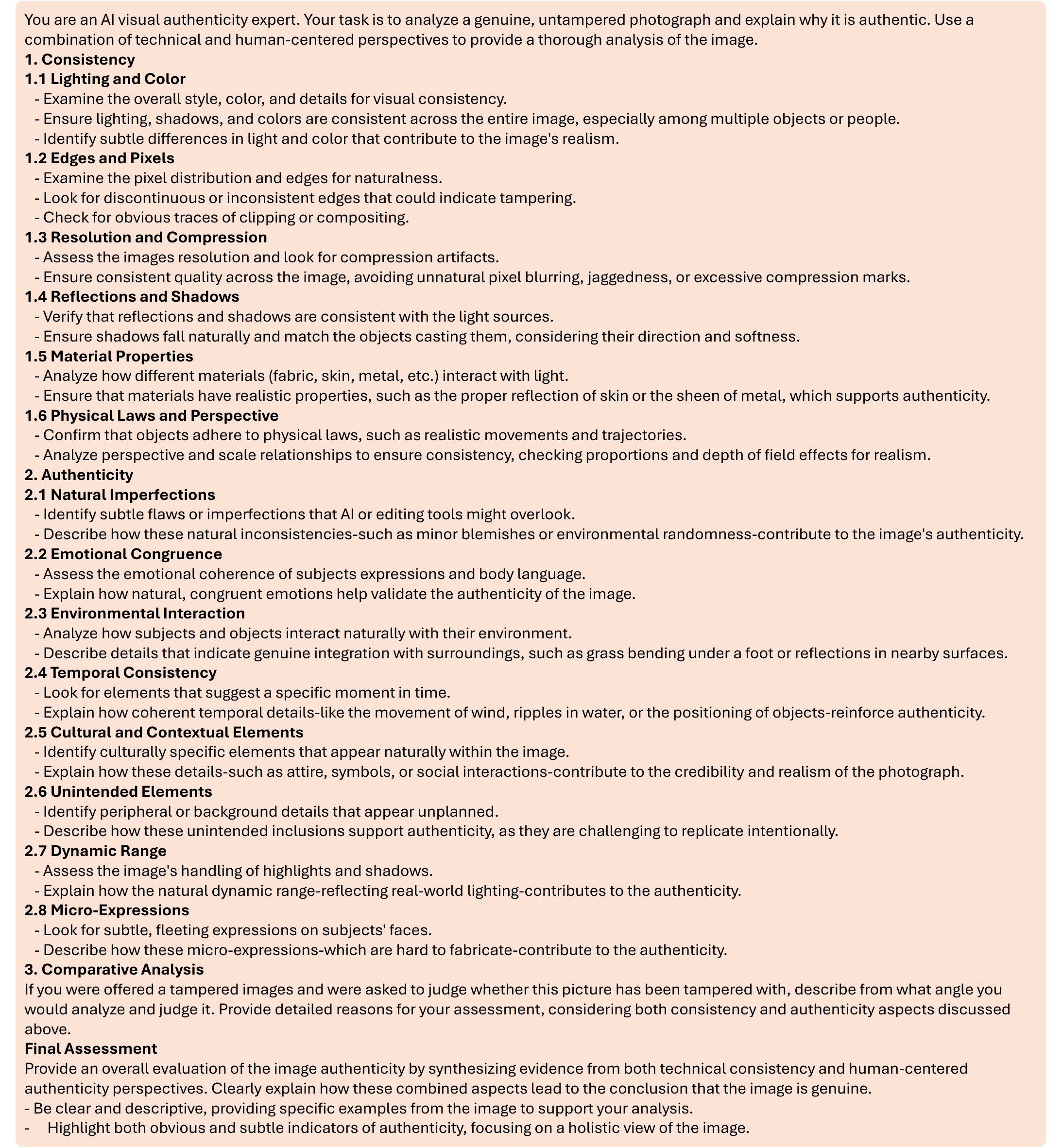}}
\caption{Prompts for real images.}
\label{suppl:Figure7} 
\end{center}
\end{figure*}

\begin{figure*}[t]
\begin{center}
\centerline{\includegraphics[width=1.8\columnwidth]{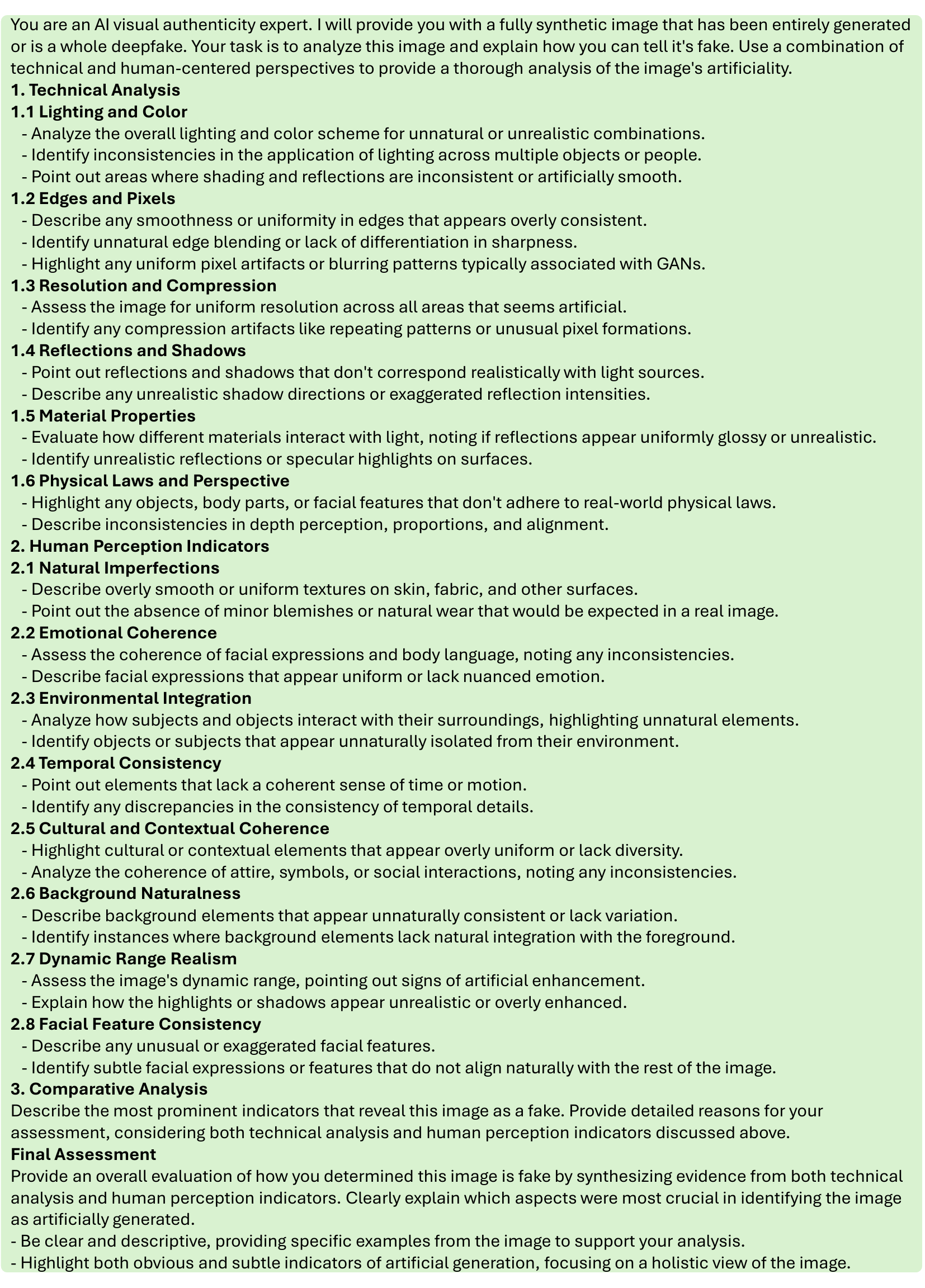}}
\caption{Prompts for fully synthetic images.}
\label{suppl:Figure8} 
\end{center}
\end{figure*}

\begin{figure*}[t]
\begin{center}
\centerline{\includegraphics[width=2.0\columnwidth]{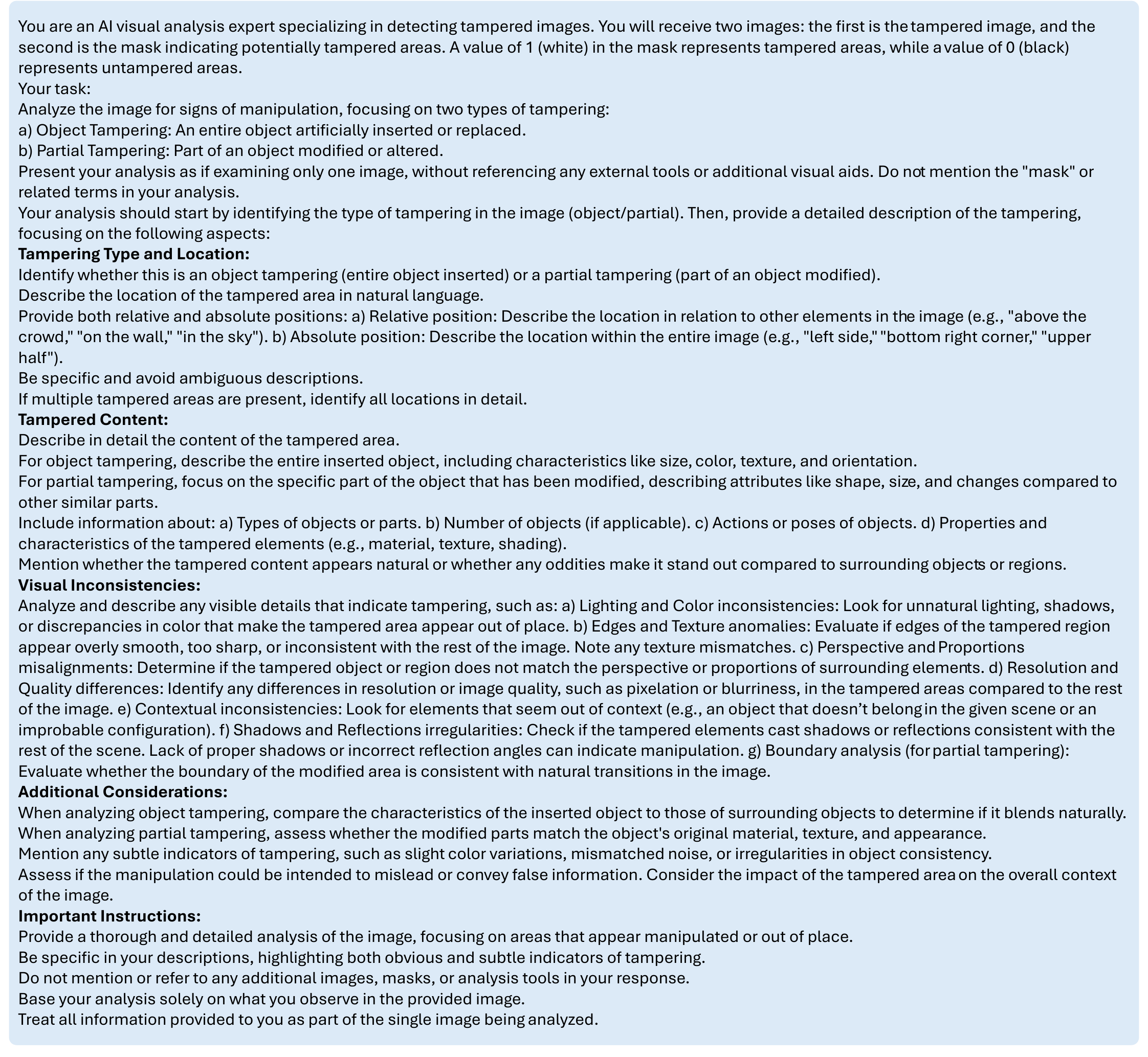}}
\caption{Prompts for tampered images.}
\label{suppl:Figure9} 
\end{center}
\end{figure*}

\begin{figure*}[t]
\begin{center}
\centerline{\includegraphics[width=2.0\columnwidth]{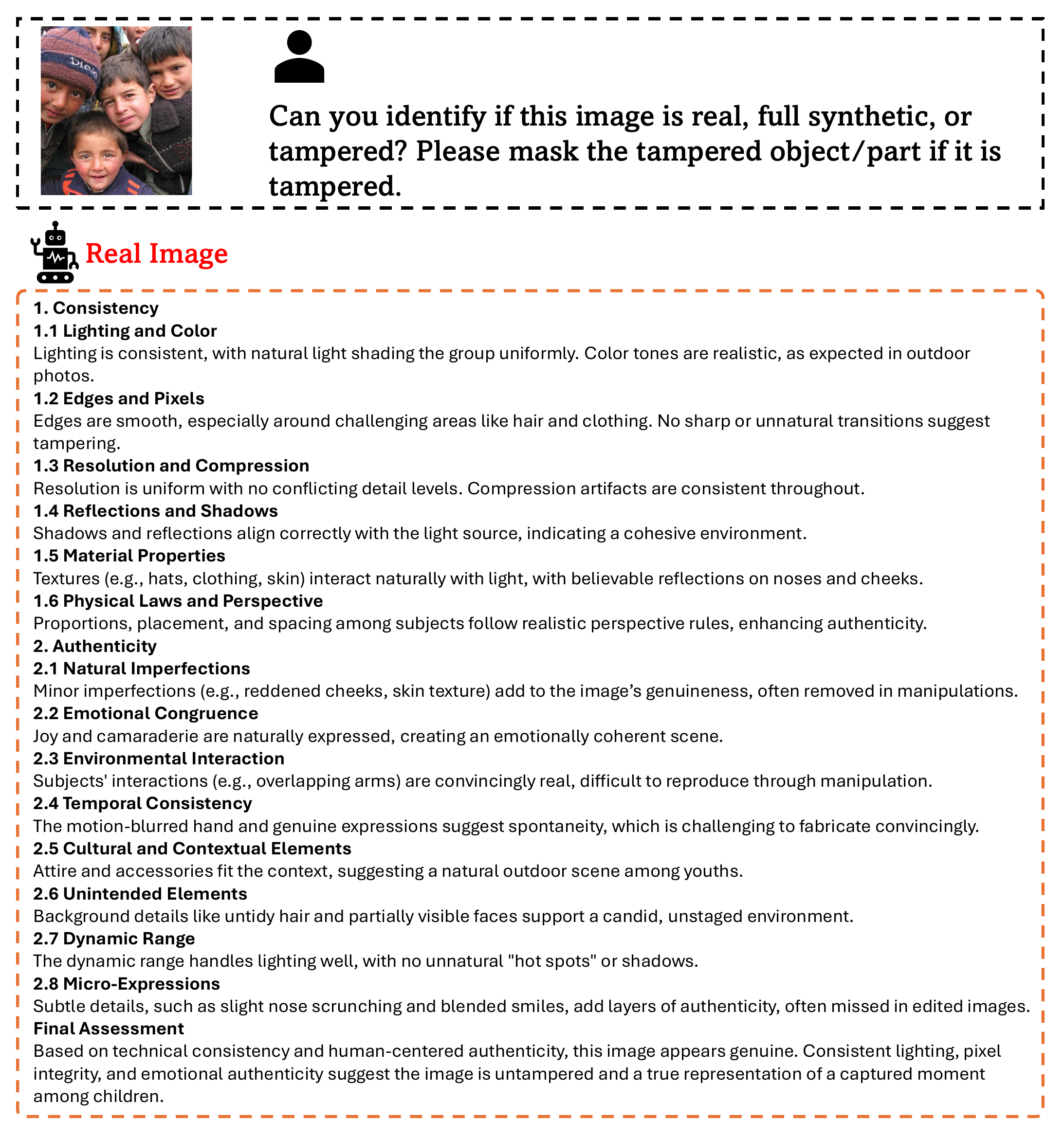}}
\caption{Example output from SIDA when analyzing a real image.}
\label{suppl:Figure10} 
\end{center}
\end{figure*}

\begin{figure*}[t]
\begin{center}
\centerline{\includegraphics[width=2.0\columnwidth]{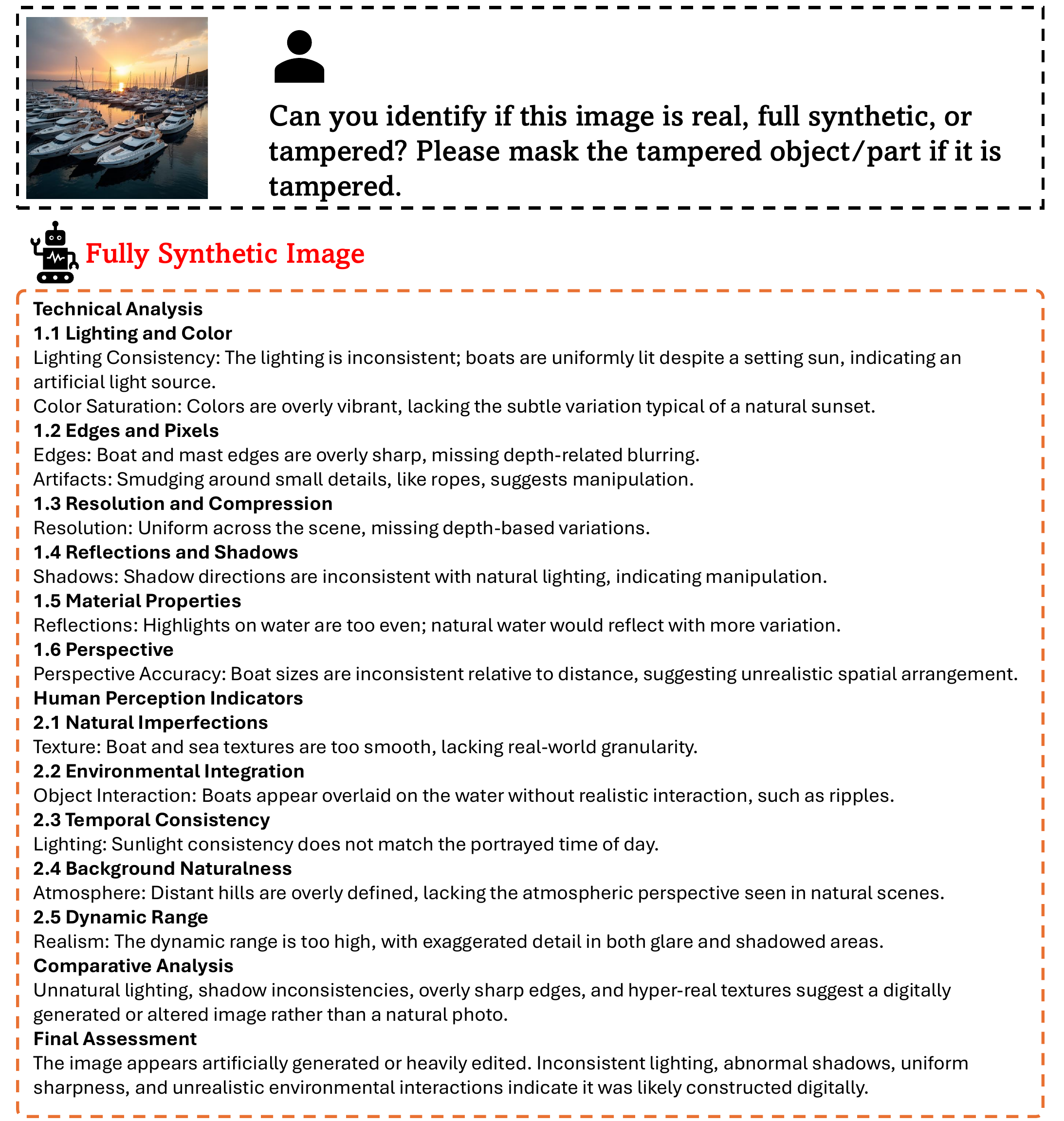}}
\caption{Example output from SIDA when analyzing a fully synthetic image.}
\label{suppl:Figure11} 
\end{center}
\end{figure*}

\begin{figure*}[t]
\begin{center}
\centerline{\includegraphics[width=2.0\columnwidth]{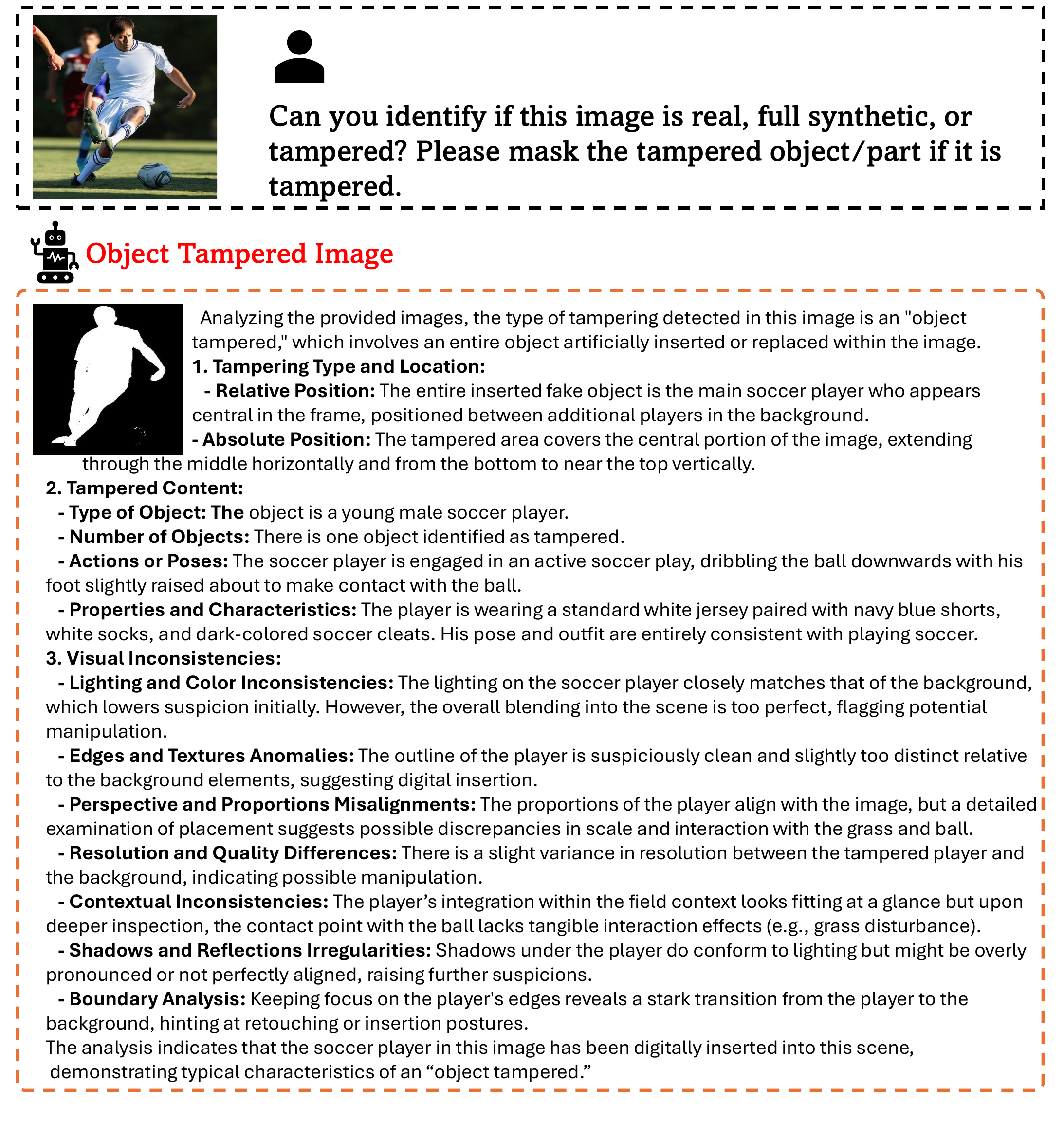}}
\caption{Example output from SIDA when analyzing a tampered image.}
\label{suppl:Figure12} 
\end{center}
\end{figure*}

\begin{figure*}[t]
\begin{center}
\centerline{\includegraphics[width=2.1\columnwidth]{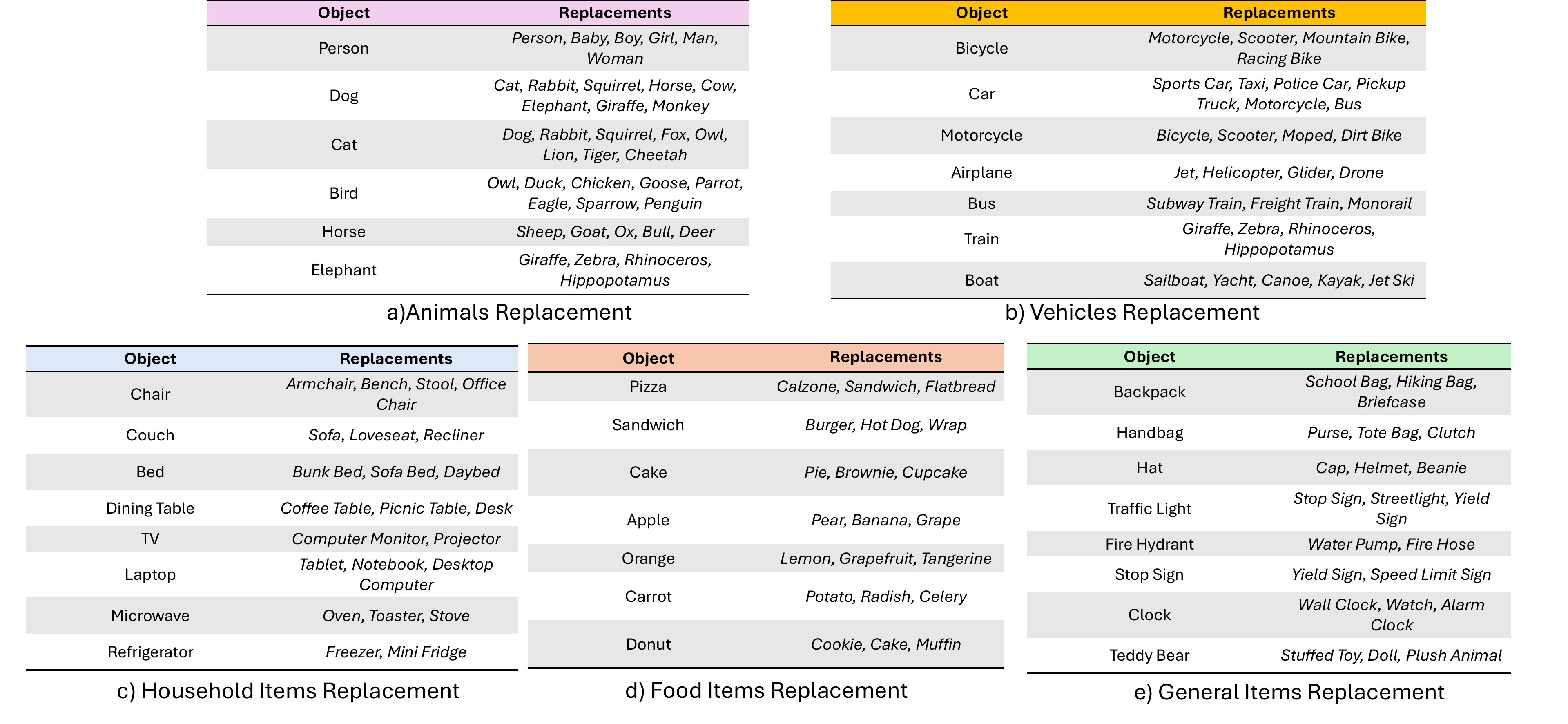}}
\caption{Object replacement directories for SID-Set.}
\label{suppl:Figure13} 
\end{center}
\end{figure*}

\begin{figure*}[b]
\begin{center}
\centerline{\includegraphics[width=2.1\columnwidth]{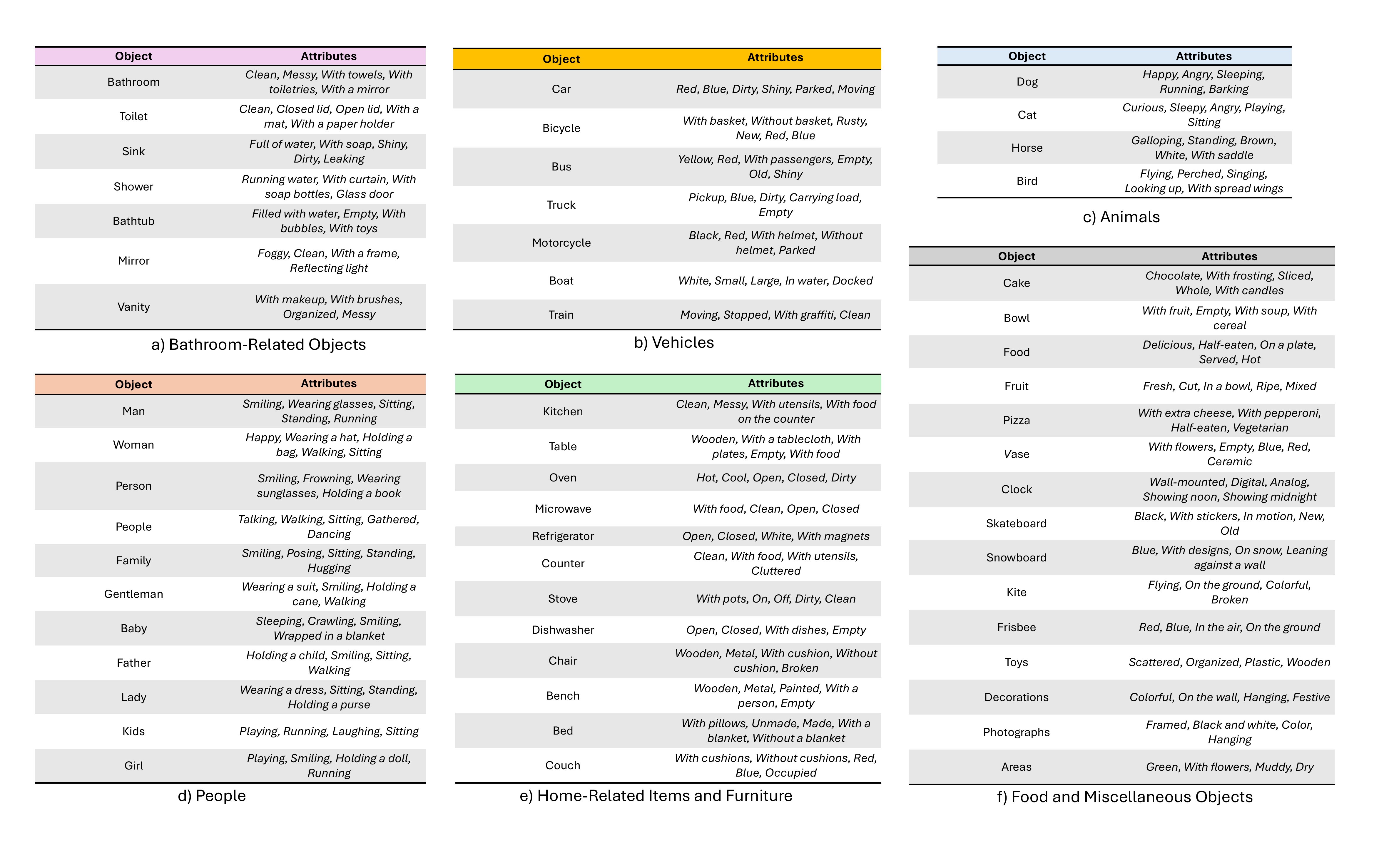}}
\caption{Attribute modification directories for SID-Set.}
\label{suppl:Figure14} 
\end{center}
\end{figure*}

\textbf{Prompts for Generating Descriptions.}
We designed prompts to generate different descriptions using GPT-4o. Separate prompts were crafted for real images, fully synthetic images, and tampered images. The prompts are illustrated in Figures~\ref{suppl:Figure7},~\ref{suppl:Figure8}, and~\ref{suppl:Figure9}.
\\
\textbf{Examples of Generated Descriptions.}
we present examples of the output descriptions generated by SIDA. Cases of real images, fully synthetic images, and tampered images are shown in Figures~\ref{suppl:Figure10},~\ref{suppl:Figure11}, and~\ref{suppl:Figure12}, respectively.
\\
\textbf{Details of Generative Process.}
We provide further details on the generation of fully synthetic and tampered images.  
\\
\textbf{\textit{Fully Synthetic Images.}}
We used FLUX\footnote{\url{https://github.com/black-forest-labs/flux}} to generate fully synthetic images due to its high quality, utilizing original data from Flickr30k~\cite{DBLP:journals/ijcv/PlummerWCCHL17} and COCO~\cite{DBLP:conf/eccv/LinMBHPRDZ14}. The style prompt was set as ``cinematic photo of {prompt}, 35mm photograph, film, professional, 4k, highly detailed," while the negative prompt included terms ``deformed iris, deformed pupils, semi-realistic, cgi, 3d, render,  sketch, cartoon, drawing, anime, text, cropped, out of frame,   worst quality, low quality, jpeg artifacts, ugly, duplicate, morbid, mutilated, extra fingers, mutated hands, poorly drawn hands, poorly drawn face, mutation, deformed, blurry, dehydrated,  bad anatomy, bad proportions, extra limbs, cloned face, disfigured, gross proportions, malformed limbs, missing arms, missing legs, extra arms, extra legs, fused fingers, too many fingers, long neck," to avoid unrealistic artifacts. All images were generated using 2 NVIDIA A100 GPUs with 40GB memory.
\\
\textbf{\textit{Tampered Images.}}
We detail each step of the tampered image generation process. Two separate directories were set up: one for object replacement, where entire objects (e.g., animals, vehicles, household items) are swapped with similar classes to generate new scenarios, and another for attribute replacement, which modifies specific features or characteristics of objects (e.g., changing an animal's emotion or activity, such as making a ``dog" appear ``happy"or ``running"). Figures~\ref{suppl:Figure13} and~\ref{suppl:Figure14} illustrate the detailed directories for object and attribute replacements, respectively. 
Furthermore, we employed GPT-4o to generate ground truth descriptions for three distinct input types: (1) real images paired with prompts, (2) fully synthetic images accompanied by corresponding prompts, and (3) tampered images provided with both prompts and their associated tampered masks, as shown in Figure~\ref{suppl:Figure18}. Figure~\ref{sec3_Figure3} elaborates on each step of the tampered image generation process using these replacement strategies. To further enrich the diversity of SID-Set, we integrated image segments from Magicbrush~\cite{DBLP:conf/nips/ZhangMCSS23}. Additional information about Magicbrush is available on its project website\footnote{\url{https://osu-nlp-group.github.io/MagicBrush/}}.

\section{Experts and Human Evaluation}
\label{suppl:experts}
We engaged five experts to undertake the following three tasks:

\textbf{Model Selection:} The experts examined approximately 1,000 images to assess the consistency and quality of the generated outputs. Based on their evaluations, we selected FLUX and latent-diffusion as our default generative models due to their superior performance.

\textbf{Image Quality Assessment:} Following image generation, the experts evaluated the realism of the outputs. To standardize this process, we introduced a five-point rating scale for image realism, ranging from 0 (lowest quality) to 5 (highest quality). Images scoring below 3 were flagged as unnatural or defective. These flagged images underwent a secondary review by the experts, after which all identified flawed images were excluded to maintain the dataset’s overall quality.

\textbf{Textual Description Evaluation:} The experts systematically reviewed 3,000 textual descriptions produced by GPT-4 to verify their semantic accuracy and alignment with the corresponding images. This evaluation adhered to three key criteria: (1) Accuracy – ensuring the description accurately reflects the image’s visual content; (2) Clarity – confirming the description is concise, unambiguous, and easily comprehensible; and (3) Consistency – verifying coherence with similar prompts or scenarios across the dataset.

These refinements have been incorporated into our revised approach. We appreciate the constructive feedback and valuable guidance provided, which have significantly strengthened this process.

\clearpage
\section*{Acknowledgments}
This work is supported by The Alan Turing Institute (UK) through the project `Turing-DSO Labs Singapore Collaboration' (SDCfP2\textbackslash 100009).
{\small
\bibliographystyle{ieeenat_fullname}
\bibliography{main}
}

\end{document}